%% file: iclr2026_conference.tex
\documentclass{article} 
\usepackage{iclr2026_conference,times}
\usepackage{graphicx}
\usepackage[table,xcdraw]{xcolor}
\usepackage{colortbl} 

\usepackage[most]{tcolorbox} 
\usepackage{enumitem}        
\usepackage{bigfoot} 
\tcbuselibrary{skins, breakable, theorems, listings, hooks}
\usepackage[table,xcdraw,dvipsnames]{xcolor}
\usepackage{marvosym}
\input{math_commands.tex}

\usepackage{hyperref}
\usepackage{url}
\usepackage{float}
\usepackage{pifont}
\usepackage{multirow}
\usepackage{booktabs}
\usepackage{newtxtext}
\usepackage{bm}
\usepackage{float}
\usepackage{subfigure}
\usepackage{multicol}
\usepackage{pdfpages}
\usepackage{indentfirst}
\usepackage{diagbox}
\usepackage{amsmath}
\usepackage{amsthm}
\usepackage{newfloat}
\usepackage{listings}
\usepackage{adjustbox}
\usepackage{booktabs}
\usepackage{caption}
\usepackage{CJK}
\definecolor{mylightblue}{RGB}{100,149,200}
\makeatletter
\def\blfootnote{\xdef\@thefnmark{}\@footnotetext}
\makeatother

\renewcommand{\cite}[1]{\citep{#1}}

\renewcommand\arraystretch{1.1}

\newcommand{\mymidrule}{
    \noalign{\vspace{-0.8mm}}
    \midrule
    \noalign{\vspace{-1mm}}
}

\title{MetaCaptioner: Towards Generalist Visual Captioning with Open-source Suites }

\author{
    Zhenxin Lei$^{4,1}$, ~Zhangwei Gao$^{2,1}$, ~Changyao Tian$^{6,1}$, ~Erfei Cui$^{2,1}$, Guanzhou Chen~$^{2,1}$, \\
    \textbf{Danni Yang$^{3,1}$, ~Yuchen Duan$^{6,1}$, ~Zhaokai Wang$^{2,1}$, ~Wenhao Li$^{5,1}$, Weiyun Wang~$^{3,1}$,} \\
    \textbf{Xiangyu Zhao$^{2,1}$, ~Jiayi Ji$^{5}$, Yu Qiao$^{1}$, ~Wenhai Wang$^{6,1}$, ~Gen Luo$^{1}$}\textsuperscript{\Letter}\\ 
    ~$^1$Shanghai AI Laboratory \quad $^2$Shanghai Jiao Tong University  \quad $^3$Fudan University \\
    ~$^4$University of Chinese Academy of Science \quad $^5$Xiamen University\\
    ~$^6$The Chinese University of Hong Kong  \\  
    [3mm]
    \textbf{~Github:}~~\textcolor{mylightblue}{\small \url{https://github.com/OpenGVLab/MetaCaptioner}} \\
}

\iclrfinalcopy 
\begin{document}

\blfootnote{\textsuperscript{\Letter}Corresponding author.}

\maketitle

\begin{abstract}
Generalist visual captioning goes beyond a simple appearance description task, but requires integrating a series of visual cues into a caption and handling various visual domains.  In this task, current open-source models present a large performance gap with commercial ones, which limits various applications such as data synthesis.  To bridge the gap,  this paper proposes CapFlow, a novel multi-agent collaboration workflow.  CapFlow demonstrates for the first time that, by capitalizing on open-source models, it is possible to achieve caption quality on par with GPT-4.1 in various domains with an 89.5\% reduction in costs.   By leveraging CapFlow as the data synthesizer, we produce high-quality visual captions from image and video domains at scale, and obtain a generalist visual captioner via fine-tuning, namely MetaCaptioner. Through extensive experiments, we show that MetaCaptioner not only achieves comparable captioning capabilities with commercial models but also reaches top-tier multimodal performance in the open-source community. We hope CapFlow and MetaCaptioner can benefit future multimodal research by providing a strong and cost-effective visual captioning solution. Our source code and models are released at ~~\textcolor{mylightblue}{\small \url{https://github.com/OpenGVLab/MetaCaptioner}}.

\end{abstract}

\begin{figure}[h]
    \centering
    \includegraphics[width=0.97\linewidth]{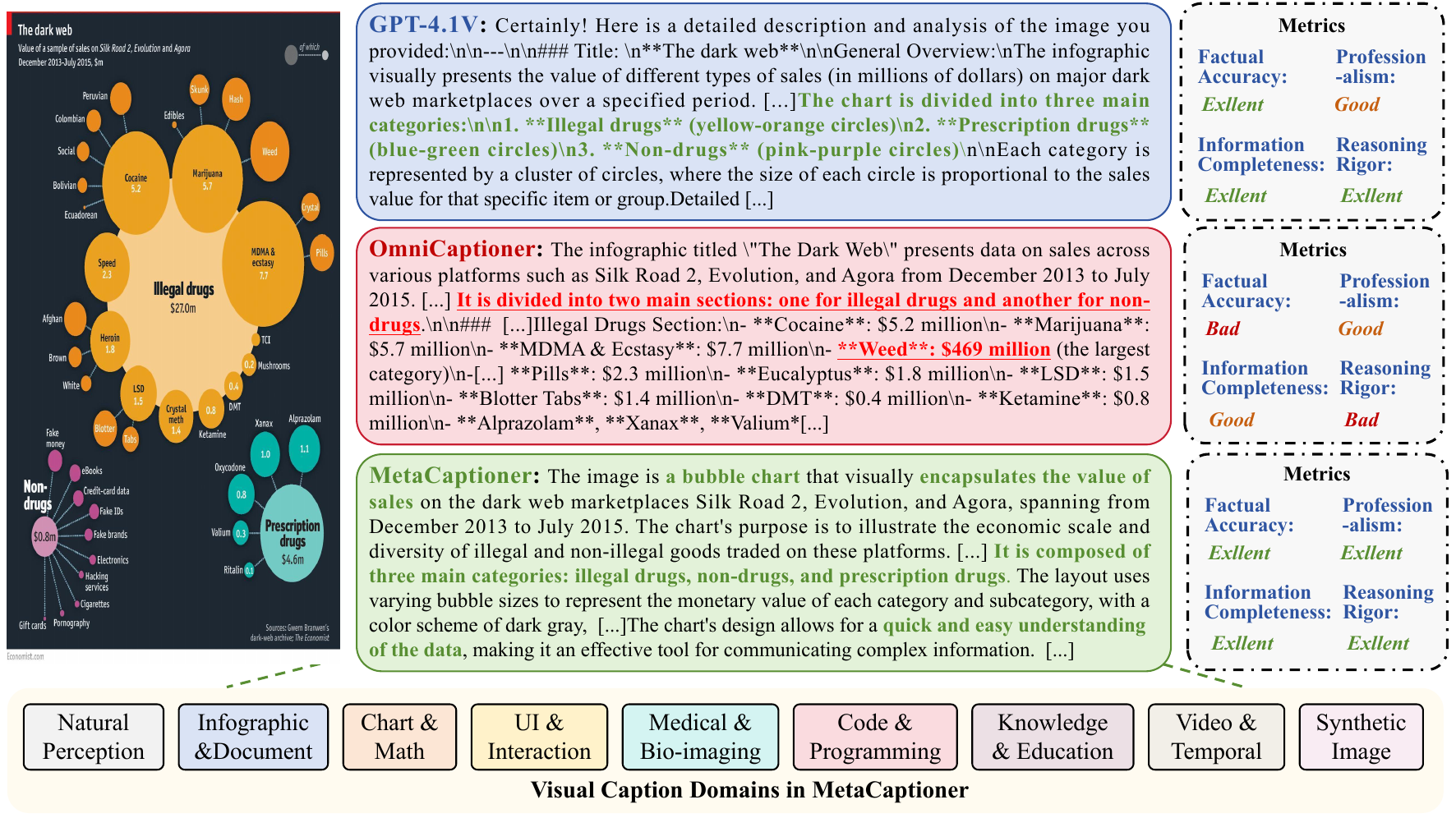}
    \vspace{-0.8em}
    \caption{\textbf{Comparison of caption quality between GPT-4.1, OmniCaptioner~\citep{Omnicaptioner} and MetaCaptioner.} These metrics are computed by GPT-5. MetaCaptioner is able to generate informative captions for a variety of visual domains with comparable quality to GPT-4.1.
    }
    \label{fig:teaser}
\end{figure}

\section{Introduction}

Visual captioning, a fundamental task in computer vision,  aims to describe images or videos using natural sentences~\citep{wang2020overview, xu2015show}. Over the past decade,  this field has received an influx of interest in the community~\citep{sharegpt4v, densefusion-1M,you2016image,anderson2018bottom}, with the majority of research focusing on natural scene description~\citep{you2016image,yao2017boosting,anderson2018bottom}. With the advent of multimodal large language models (MLLMs)~\citep{ Internvl3_5, qwen2-vl,  llavaOV}, visual captioning has evolved beyond a simple multimodal task and now plays an increasingly crucial role in model pre-training~\citep{Internvl3_5, dalle3} and data synthesis~\citep{sharegpt4v,  densefusion-1M, Capsfusion}. This progression imposes higher demands on existing visual captioners in handling more diverse and complex scenarios, such as mathematical figures~\citep{MathVerse, MathVista}, diagrams~\citep{InfoVQA}, and documents~\citep{DocVQA}.

To overcome this limitation, increasing efforts have been devoted to producing informative visual captions. Among them, a promising method is to design domain-specific prompt and leverage the advanced commercial MLLM, \emph{e.g.,} GPT-4~\citep{gpt4o}, to generate detailed captions~\citep{sharegpt4v}.  While effective, such approach becomes prohibitively expensive as the data scale increases.  Recent attempts also explore open-source MLLMs to synthesize the generalist visual captions~\citep{scalecap,Omnicaptioner}, yet their caption quality still lags far behind  that of commercial ones due to the limited capabilities. Therefore, one critical issues are naturally raised: \textit{Is possible to bridge the gap in generalist visual captioning through open-source suites}?

To answer this question, we observe that the main difficulty of generalist visual captioning with open-source MLLM lies in the large domain gap. In particular, most MLLMs excel at natural scenes, and they tend to describe the subject, texture, and object relationship in an image. Nevertheless, the description requirements become quite different and challenging in other visual domains. For instance, for knowledge-based scenario, \emph{e.g.,} math, the description  should not only contain visual appearance, but also the related visual knowledge and underlying visual logic. As shown in Fig.~\ref{fig:teaser}, open-source MLLMs like OmniCaptioner~\cite{Omnicaptioner} struggle to seamlessly integrate these capabilities into one description,  often lacking fine-grained details in some aspects.

\begin{figure}
    \centering
    \includegraphics[width=\linewidth]{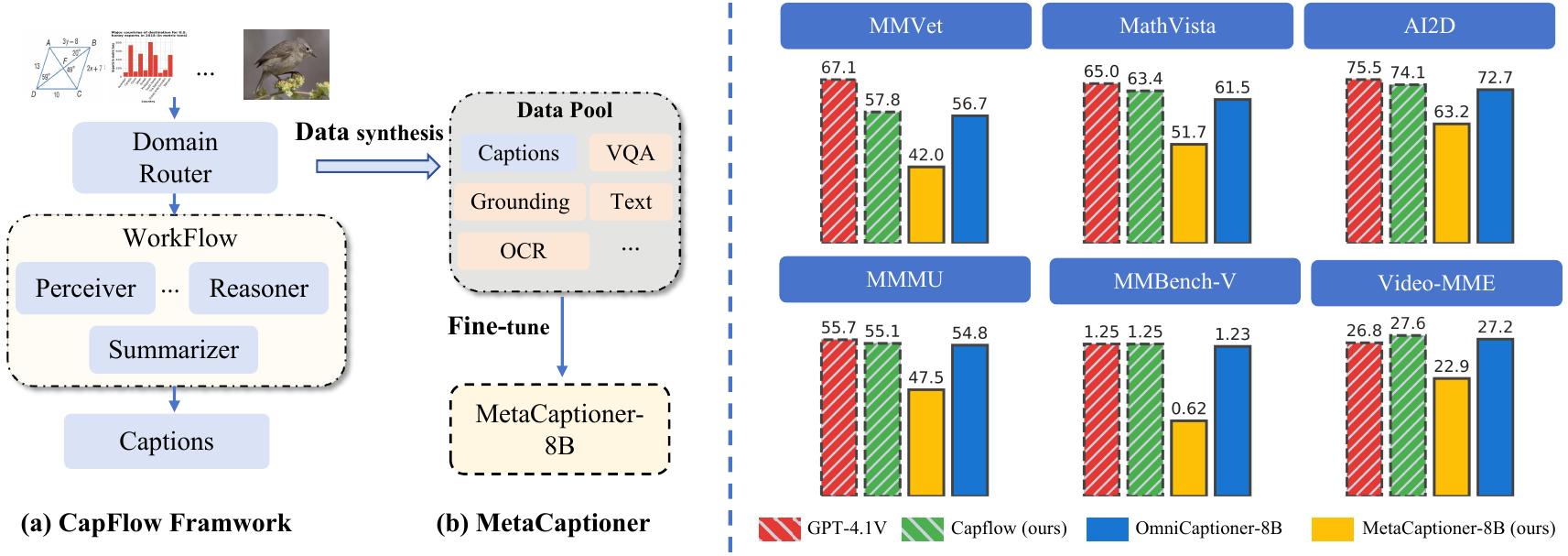}
    \vspace{-0.8em}
    \caption{\textbf{Comparison of existing captioning systems with our our methods.} Our CapFlow adopts a multi-agent collaboration workflow to produce high-quality captions for training MetaCaptioner-8B.   Under the setting of visual reasoning with LLMs, CapFlow and MetaCaptioner-8B achieve comparable  performance with GPT-4.1 using open-source suites.   }
    \label{fig:comparison}
        \vspace{-1em}
\end{figure}

To address this issue, we propose \textit{CapFlow}, a novel multi-agent collaboration workflow for producing generalist visual captioning with open-source MLLMs.  The main principle of CapFlow is to decompose visual captioning into sub-tasks and synthesize evidence from diverse aspects.   As shown in Fig.~\ref{fig:comparison}, CapFlow consists of a hierarchical workflow, where each agent performs as a distinct role in perception, visual knowledge extraction,  visual reasoning, summary, \textit{etc}.  In CapFlow, visual descriptions from different perspectives are produced by the corresponding agents and ultimately aggregated into a single caption.  To further eliminate the domain gap, we design a domain routing mechanism to dynamically assign the suitable workflow of CapFlow for different visual domains.  Through close collaboration, CapFlow can achieve general visual captioning capabilities comparable to the top commercial models \emph{i.e.} GPT-4.1~\citep{gpt4.1}, at only 10.5\% of the cost.

Based on CapFlow, we aim to equip generalist visual captioning capabilities to common MLLMs~\citep{Internvl3_5}, thus yielding an  efficient and general visual captioner, namely \textit{MetaCaptioner}.  In particular, we leverage CapFlow as a strong data engine to produce informative captions for images and videos from various domains, with a strict reject sampling pipeline to filter out low-quality captions.  Thanks to the cheap yet effective pipeline of CapFlow, we can efficiently scale up the data size to 4.1 millions, enabling a seamless convert of the open-source MLLM into MetaCaptioner through fine-tuning.   Thanks to the high-quality captions from CapFlow, MetaCaptioner not only demonstrates powerful visual captioning capabilities at a lower cost (  0.7\% of GPT-4.1), but also achieves top-tier multimodal performance against existing open-source MLLMs.

To validate our approach, we conduct extensive experiments on 13 benchmarks of two common settings: (1) Visual reasoning with large language models (LLMs)~\citep{Omnicaptioner}: The detailed captions generated by CapFlow and MetaCaptioner are used as input to LLM for visual reasoning, thus diagnosing visual caption capabilities. (2)  Downstream evaluations~\citep{Internvl3_5,qwen2-vl,xie2025sana}: We directly evaluate the performance of MetaCaptioner on understanding and reasoning  tasks across various domains, thereby validating the benefits of training with generalist caption data.  Our experimental results show that our MetaCaptioner, even with a lightweight parameter size, can already achieve comparable captioning capabilities with the commercial model GPT-4.1~\citep{gpt4.1} on several benchmarks, and its downstream performance also benefits from the synthetic data from CapFlow, \emph{e.g.,} +3.7 on MathVerse. CapFlow and MetaCaptioner provide a cheap and strong visual captioning solution that will potentially benefit future multimodal research.  In summary, our contributions are three folds:
\begin{itemize}[leftmargin=*,itemsep=1pt, topsep=0pt, parsep=0pt]
    \item We propose CapFlow, an innovative multi-agent collaboration workflow for producing generalist captioning on various image and video domains.  CapFlow is the first open-source  framework that  reaches comparable captioning performance with   GPT-4.1.
    \item By leveraging CapFlow as the data synthesizer, we present the strongest open-source visual captioner in the community, namely MetaCaptioner. MetaCaptioner not only possess powerful  generalist captioning capabilities, but also maintains top-tier multimodal performance.
    \item We conduct extensive experiments to ablate CapFlow and MetaCaptioner, providing invaluable hints for future research. In addition, our source models and codes will be publicly released, further advancing  the development of generalist visual captioning and the broader multimodal research. 
\end{itemize}

\section{Related Work}

\subsection{Multimodal Large Language Models}

With the development of multi-modal large language models (MLLMs), visual-language alignment has attached more attention. Most approaches improve the alignment of visual and language by pre-training on large-scale image-text pairs \citep{CLIP, Flamingo, Blip, Blip2}. Subsequent research attempt to achieve this challenge through incorporating high quality data for pre-training and supervised fine-tuning stage \citep{LLaVA, llavaOV, Llava-uhd, Minigpt-4, qwen2-vl, InternVL2.5, Internvl3_5}. Some models further introduce more fine-grained image-text annotation data\citep{densefusion-1M, DenseWorld-1M, Coconut-pancap}(\emph{e.g.} grounding and mask annotations) to reduce hallucinations and strengthen visual grounding ability. Nevertheless, due to the domain coverage and annotation costs, constructing multi-domain, high-quality image-caption pairs remains a significant challenge.

\subsection{Generalist Visual Captioning}

Recent progress in multimodal community has significantly increased the demand for generalist visual captioning. In particular, existing  methods can be broadly categorized into two categories. The first category often prompts powerful MLLMs (\emph{e.g.}, GPT-4o \citep{gpt4o} and GPT-4V \citep{gpt4} through carefully designed prompts to generate high-quality captions~\citep{ sharegpt4v, sharegpt4video, Omnicaptioner}. Some works also adopt coarse-to-fine grained optimization 
 like introducing visually differentiated descriptions \citep{sharegpt4video} and multi-turn caption optimization \citep{dalle3, scalecap, Pix2Cap-COCO}. The second category centers on multi-source caption synthesis, typically incorporating multiple domain expert models (\emph{e.g.}, SAM\citep{sam}, Trace-Uni\citep{traceuni}, and Paddle OCR\citep{PaddleOCR}) and merging the domain expert results \citep{densefusion-1M, DenseWorld-1M, CAT, DAM, Coconut-pancap}.  However, both categories face limitations in content granularity and comprehension depth, often exceeding the capabilities of open-source MLLMs.  Therefore,  approaching generalist captioning with open-source models remains a significant challenge in the community.

\section{CapFlow}

\begin{figure}[t]
    \centering
    \includegraphics[width=\linewidth]{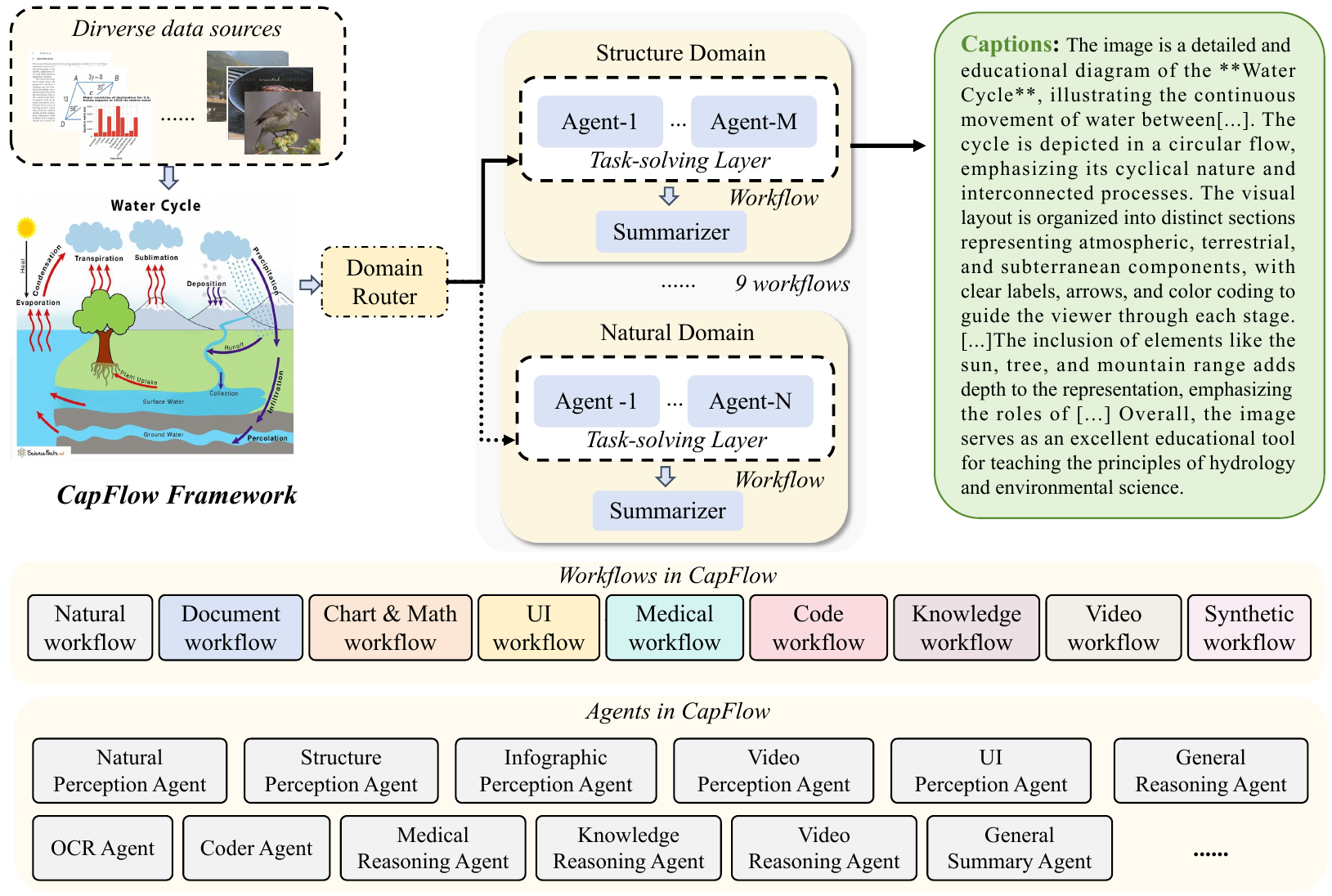}
    \vspace{-1.5em}
    \caption{\textbf{Overview of CapFlow Framework.} CapFlow can dynamically select an appropriate caption workflow for the input image, and each workflow is equipped with multiple  agents for close collaboration.  With this pipeline, CapFlow can produce high-quality captions of various visual domains with open-source MLLMs, \emph{e.g.,} Qwen2.5-VL \citep{qwen2_5_vl}.  
    }
        \vspace{-1em}
    \label{fig:overview}
\end{figure}
\vspace{-0.5em}

\subsection{Overview}

A core challenge in generating diverse visual descriptions across different domains lies in the varying requirements for granularity and semantic depth \citep{scalecap, Omnicaptioner}. In this work, we propose CapFlow, a novel multi-agent collaboration framework for automated visual description. As illustrated in Fig. \ref{fig:overview}, CapFlow consists of two main parts: a \textit{Domain Routing} and  a \textit{Hierarchical Captioning Workflow}. The design of CapFlow features in: 1) high-quality caption generation that integrates various visual cues, 2) collaborative agent framework applicable to a wide range of visual domains, and 3) cheap deployment cost and high scalability.

\subsection{Domain Routing}
The motivation for the domain routing originates from an intuitive principle: \textit{various visual domains demand varying levels of cognitive complexity}~\citep{Zongshu1O1, Zongshu2}. Therefore, CapFlow needs to first determine the domain of the visual input and then assign it a workflow suitable for visual captioning. To approach this target, we first adopt an MLLM as a domain router, with a carefully-designed prompt to query the visual domain of images. This process is similar to solving a visual question-answering problem, where the MLLM should select the best option from our pre-defined domain based on the visual input and the prompt. For data with a known domain, we can skip the domain routing part.

\subsection{Hierarchical Captioning Workflows} Based on the routing results, we design  different routing paths for visual captioning of corresponding domains, and each routing path is customized for a visual domain.     As shown in Fig.~\ref{fig:overview}, the routing path is designed as a  hierarchical  workflow,  with two distinct layers for task solving and information summarization, respectively.

In particular, the task-solving layer aims to decompose a complex captioning task into several simple sub-tasks and perform divide-and-conquer through multiple agents.  Each domain is equipped with a distinct set of agents to solve domain-specific tasks, as shown in Tab.~\ref{agentworkflow}. Therefore, we design a variety of functional agents that are responsible for different tasks, where most of the agents are based on LLMs. In particular, there are a total of 30 functional agents in the workflow, which can be roughly divided into four categories: \textit{guideline, perception, reasoning, and tools.} Specifically, the guideline agents aim to provide an overview description of the visual input, mainly containing global information like style, structure, and geography. In contrast, the perception agents depict the fine-grained details of images and videos such as textures, colors, and lighting. These two types of agents mainly focus on visual appearance. To compensate for high-level semantics, the reasoning agent is responsible for capturing the underlying visual knowledge, logic, and relationships. In addition, we notice that some domains often require additional visual tools to accommodate their requirements. For example, the document-related data often requires OCR parsers to obtain textual words. Therefore, we also design the tool agent to perform as specific tools like the OCR-based parser and the code parser. The detailed prompts are provided in the Appendix.

The visual cues generated by the task-solving layer  are ultimately aggregated into one caption through the  information summarization layer. In this layer, an LLM agent aims to summarize the textual contents from the  task-solving layer one-by-one, analyze their contributions to the final captions, and organize into an informative, detailed and structured caption.

\begin{figure}[t]
    \centering
    \includegraphics[width=\linewidth]{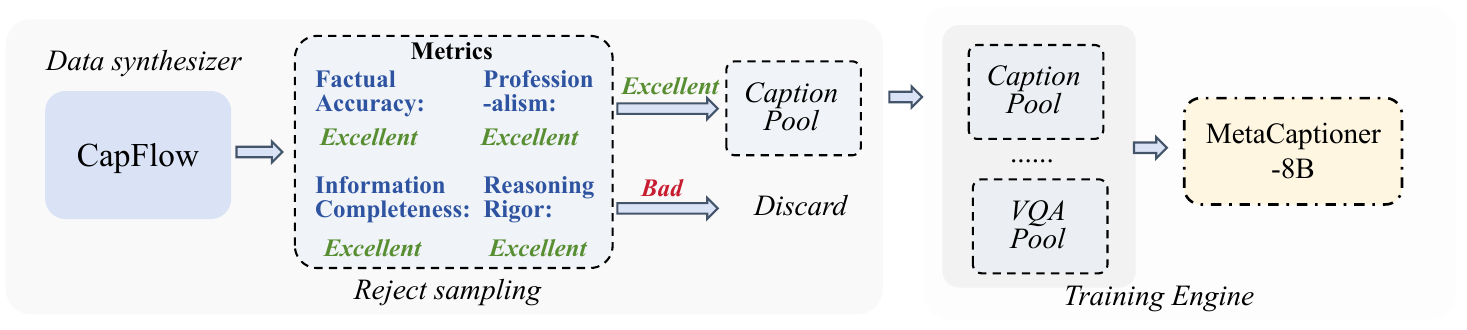}
    \vspace{-1.5em}
    \caption{\textbf{Training pipeline of MetaCaptioner.} We leverage CapFLow as the data synthesizer to generate various captions, upon which a strict reject sampling pipeline is used to filter out low-quality ones. The obtained captions are combined with common instruction data to train MetaCaptioner.  
    }
    \label{fig:training}
        \vspace{-1em}
\end{figure}

\section{MetaCaptioner}
Based on Capflow, we aim to train a strong generalist captioning model. To this end, we first collect images and videos from various domains, then employ Capflow as the data synthesizer to generate massive high-quality captions, and ultimately obtain MetaCaptioner-8B through fine-tuning.
%

\subsection{Data Collection}

\noindent \textbf{Data Source and Processing.} We first collect approximately 70 million raw images and videos from 140+ open-source image and video datasets. These sources span a wide range of domains, including natural images (\emph{e.g.,} SAM~\citep{sam} and LAION~\citep{laion}), structured images (\emph{e.g.,} ArxivQA~\citep{Arxivqa} and NovaChart~\citep{novachart}), knowledge-based images (\emph{e.g.,} PMC-VQA~\citep{PMC-VQA} and iNat~\citep{inat}), and aesthetic images (\emph{e.g.,} PixArt~\citep{Pixart-sigma}).  Due to the imbalance in different domains, we further collect a large mount of structured images (\emph{e.g.,} charts, posters, GUI screenshots), text recognition images (\emph{e.g.,} documents, reports), and commonsense images (\emph{e.g.,} natural science, human activities). Following the protocol of \citep{Internvl3}, we process these open-source datasets and obtain approximately 18 million structured images. To ensure the visual quality, we apply filtering based on image resolution and aspect ratio \citep{scalecap, densefusion-1M, DenseWorld-1M}, retaining images with a short edge larger than 512 pixels, an aspect ratio less than 2, and videos with a resolution higher than 480p. As a result, we construct a high-quality dataset containing 5 million semantically rich and diverse samples from a broad range of sources. More details  are provided in the Appendix. 

\noindent \textbf{CapFlow as the Data Engine.}

To accommodate data from diverse domains, we adopt CapFlow as the data engine for annotating the collected data to construct the dataset called MetaCaption-5M. For the functional agents, we adopt Qwen2.5-VL-72B as the functional agent and carefully design task-specific prompts to guide the agent toward fine-grained visual perception and understanding. For the summary agent, we leverage the strong contextual processing capabilities of Qwen2.5-72B to integrate and summarize information from various functional agents, ultimately generating high-quality visual captions. The prompts for dataset constructing are provided in the Appendix.

\begin{table}[t]
\renewcommand{\arraystretch}{1.1}
\centering
\resizebox{\columnwidth}{!}{%
\begin{tabular}{c|c|c}
\toprule[1.5pt]
\textbf{Visual Domain} & \textbf{Functional Agent} & \textbf{Summary Agent} \\ \mymidrule
Natural & Natural Perception, General Reasoning, Visual Guideline & \multirow{8}{*}{General Summary} \\
Structure \& Math & Structure Perception, Infographic Perception, General Reasoning, Visual Guideline &  \\
Infographic \& Document & Infographic Perception, OCR, General Reasoning, Visual Guideline &  \\
Medical \& Bio-Imaging & Natural Perception, Medical Reasoning, Visual Guideline &  \\
UI \& Interaction & UI Perception, OCR, General Reasoning &  \\
Code \& Programming & Coder, General Reasoning, Visual Guideline &  \\
Knowledge \& Education & Infographic Perception, Knowledge Reasoning, Visual Guideline &  \\
Synthetic & Texture Perception, General Reasoning, Visual Guideline &  \\ \mymidrule
Video \& Temporal & Video Perception, Video Reasoning, Video Guideline & Video Summary \\
\noalign{\vspace{-0.6mm}}
\bottomrule[1.5pt]
\end{tabular}%
}
\vspace{-0.5em}
\captionsetup{justification=justified, singlelinecheck=false}
\caption{ \textbf{The agentic workflow of various visual domains.} We decomposite the visual description task into visual guideline,  perception,  reasoning, and tool task.}
\vspace{-1em}
\label{agentworkflow}
\end{table}

\subsection{Reject Sampling Pipeline}

As illustrated in Fig.~\ref{fig:training}, we further develop a strict reject sampling strategy to filter out low-quality data from the MetaCaption-5M dataset generated by CapFlow. Specifically, we employ carefully designed prompts to ask MLLMs to score the quality of each caption from multiple aspects. For the image modality, captions are evaluated across five dimensions: \textit{Factual Accuracy}, \textit{Information Completeness}, \textit{Reasoning Rigor}, \textit{Core Intent Capture}, and \textit{Professionalism}. For the video modality, captions are assessed based on \textit{Temporal and Factual Accuracy}, \textit{Event and Detail Coverage}, \textit{Temporal Causal Logic}, \textit{Core Intent}, and \textit{Professionalism}. To minimize the influence of the model's subjective judgment on the evaluation, we adopt a 3-point rating scale for scoring the results and only sample those that achieve professional levels (\emph{e.g.}, rating 3) across all sub-domains for subsequent model training. As a result, we obtain a high-quality  dataset called MetaCaption-4.1M.

\subsection{Training Pipeline}

Based on MetaCaption-4.1M, we trained a powerful multimodal large language model,  namely MetaCaptioner-8B. Specifically, MetaCaptioner-8B adopts InternViT-600M~\cite{InternVL2.5} and Qwen3-8B-Instruct~\citep{qwen3} as the vision encoder and the language model, respectively. In the pre-training stage, we follow InternVL3.5 to perform native pre-training on various visual and text data~\cite{Internvl3_5}. For instruction tuning, we mix our generalist captioning data (MetaCaption-4.1M) with  instruction data from InternVL3.5~\cite{Internvl3_5} to optimize the entire model. Training schedule and steps are kept consistent with   InternVL3.5~\cite{Internvl3_5}.

\section{Experiment}

\subsection{Implementation Details}

In CapFlow, we utilize Qwen2.5-VL-72B~\citep{qwen2_5_vl} and Qwen2.5~\citep{qwen3} as functional agents and summary agents, respectively. The entire annotation process requires approximately 480 H200 GPU days. For reject sampling, we employ Qwen2.5-VL-7B as the judger to provide the metric score, taking around 0.1 H200 GPU days. For pre-training of MetaCaptioner, 

it is trained for one epoch with a learning rate of 1e-5 and a batch size of 256. For supervised fine-tuning (SFT), we train MetaCaptioner for 160k iterations with a learning rate of 2e-4. The complete training process takes approximately 192 H200 GPU days.

\begin{table}[]
\renewcommand{\arraystretch}{1.2}
\centering
\resizebox{\columnwidth}{!}{%
\begin{tabular}{l|cccccccc|c}
\toprule[1.5pt]
\textbf{Model} & \textbf{MMMU} & \textbf{MMVet} & \textbf{\begin{tabular}[c]{@{}c@{}}Math\\ Verse\end{tabular}} & \textbf{\begin{tabular}[c]{@{}c@{}}Math\\ Vista\end{tabular}} & \textbf{\begin{tabular}[c]{@{}c@{}}Chart\\ QA\end{tabular}} & \textbf{\begin{tabular}[c]{@{}c@{}}Info\\ VQA\end{tabular}} & \textbf{AI2D} & \textbf{\begin{tabular}[c]{@{}c@{}}Video\\ MME\end{tabular}} & \textbf{Cost} \\ \mymidrule
{GPT-4.1} & 55.7 & 61.7 & 56.8 & 65.0 & 62.3 & 63.2 & 75.5 & 26.8 & \$1.47 \\ \mymidrule
\textit{Baseline} (8B) &50.7  &47.2  &42.1  &57.6  &54.4  & 49.0 & 64.5  & 23.9 & \textbf{\$0.01} \\
+ Hierarchical Workflow & 51.6 & 48.6 & 44.0 & 59.0 & 57.8 & 44.2 & 66.0 & 26.1 & \$0.02 \\
+ Domain Routing & 54.7 & 50.5 & 43.9 & 58.7 & 58.0 & 45.0 & 67.4 & 26.2 & \$0.02 \\
\rowcolor[HTML]{EFEFEF} 
{+ {Scale up to 72B}} & 55.1  & 57.8  &53.1  &62.5  &59.2  & 50.2 &74.2  & 27.6 &  \$0.14 \\ 
\noalign{\vspace{-0.6mm}}
\bottomrule[1.5pt]
\end{tabular}%
}
\vspace{-0.5em}
\caption{\textbf{Ablation study of CapFlow.} Our baseline is Qwen2.5-VL-8B~\citep{qwen2_5_vl}. Through our careful design, CapFlow can achieve comparable performance to GPT-4.1~\citep{gpt4.1} on most benchmarks, but at a much lower cost. We calculate the average inference cost on 100 samples.
}
\vspace{-0.8em}
\label{capflow}
\end{table}

\subsection{Evaluation Benchmarks}
We evaluate our methods on 13 comprehensive multimodal benchmarks. Specifically, visual question answering benchmarks include InfographicVQA \textit{test} \citep{InfoVQA}, ChartQA \textit{test} \citep{ChartQA}, Document VQA \citep{DocVQA}, and AI2D \textit{test} \citep{AI2D}. MLLM benchmarks encompass MMBench \textit{V1.1} \citep{MMBench}, MMMU \textit{val} \citep{MMMU}, MMVet \citep{Datasets:MM-vet}, MathVista \textit{testmini} \citep{Datasets:Mathvista}, MathVerse \textit{vision only} \citep{MathVerse}, SEEDBench2 Plus \citep{SEEDBench2P}, and MMStar \citep{MMStar}. Video understanding benchmarks contain VideoMME \citep{VideoMME} and MMBench-Video \citep{MMB-V}. The evaluation metrics follow existing methods using VLMEvalkit\citep{VLMEvalkit}.

\subsection{Evaluation Settings}

\noindent \textbf{Caption Quality Evaluation.}  Similar to the metrics in reject sampling, we define five metrics that can comprehensively reflect the caption quality: factual accuracy, completeness, reasoning rigor, intent capture, and professionalism.  For these metrics, we design strict evaluation prompts  for GPT-5 to assign a score from 1 (bad) to 3 (excellent) to each metric, see Appendix.

\noindent \textbf{Visual Reasoning with LLMs.}

In this setting, visual captioners will  generate description as the visual prompt for existing reasoning-enhanced LLMs, enabling them to seamlessly perform various multimodal downstream tasks. This setting decouples visual understanding from textual reasoning, where downstream performance highly depends on  the generated captions.  Therefore, we can quantitatively diagnose the informativeness  of  captions from the  downstream performance.

\noindent \textbf{Direct Multimodal Evaluation.}

Direct multimodal evaluation is the most common setting to assess the capabilities of MLLMs on downstream tasks~\citep{Internvl3_5,qwen2-vl}. Through this setting,   we aim to diagnose the impact of high-quality synthetic data (\emph{i.e.,} \textit{MetaCaption-4.1M}) on the multimodal capabilities of MetaCaptioner.

\subsection{Ablation Study}

\subsubsection{Ablation Study of CapFlow}
In Tab.~\ref{capflow}, we ablate the design of CapFlow and compare it with GPT-4.1~\citep{gpt4.1} under the setting of \textit{visual reasoning with LLMs}.  In this table, we first observe that starting from the Qwen2.5-VL-8B baseline, the introduction of the Hierarchical Workflow brings moderate gains on most benchmarks, \textit{e.g.,}+1.4 on MMVet and +1.9 on MathVerse. The subsequent integration of the Domain Routing to categorize different domains yields more pronounced improvements, particularly on MMMU (+4.0 over baseline), MMVet (+3.3 gains), indicating its critical role in addressing domain gaps. Notably, scaling the base model of CapFlow to 72B parameters leads to substantial performance boosts across all tasks, nearing GPT-4.1 on several fronts: MMMU (55.1 vs. 55.7), MathVista (62.5 vs. 65.0), and AI2D (74.2 vs. 75.5). On VideoMME, CapFlow-72B even exceeds GPT-4.1 (27.6 vs. 26.8), highlighting its strong capabilities in video domains.  
Most importantly, these results are achieved at only 10.5\% of the cost of GPT-4.1 solutions, \textit{i.e.,} \$0.14  \textit{vs.} \$1.5 per image, making high-quality caption synthesis feasible for large-scale applications.
In summary, the ablation study confirms that both the workflow design and model scaling are essential to the performance of generalist captioning.

\subsubsection{Impact of Synthetic Captions on Training}
In Tab.~\ref{pretrain&sft}, we systematically evaluate the impact of integrating different types of synthetic caption data~\citep{sharegpt4v,densefusion-1M} on pre-training and fine-tuning.    As shown in  Tab.~\ref{pretrain&sft}, when augmenting pre-training data, CapFlow demonstrates clear advantages in several key benchmarks, with  the highest scores on MMVet (62.5), InfoVQA (68.3), MMMU (56.1), and AI2D (77.5).  Considering the very competitive baselines, these gains confirm the benefits of captions generated by CapFlow to MLLM pre-training.  
In the fine-tuning phase, incorporating CapFlow data also leads to notable improvements, particularly on MMVet (62.8) and MathVista (60.9), where it achieves the best average performance among all configurations.  Notably, Video-MME results (61.7 for MetaCaption) highlight its competence in temporal visual reasoning, outperforming other synthetic data.   In conclusion, the synthetic captions produced by CapFlow consistently enhance model performance across both training stages, demonstrating their richness, accuracy, and broad applicability.

\subsection{Main Results}

\begin{table}[!]
\setlength{\tabcolsep}{3pt}
\renewcommand{\arraystretch}{1.2}
\centering
\resizebox{\columnwidth}{!}{%
\begin{tabular}{l|c|l|cccccccc|c}
\toprule[1.5pt]
\textbf{Model} &
  \textbf{\begin{tabular}[c]{@{}c@{}}Pre-training\\ Data\end{tabular}} &
  \multicolumn{1}{c|}{\textbf{\begin{tabular}[c]{@{}c@{}}SFT\\ Data\end{tabular}}} &
  \textbf{MMMU} &
  \textbf{MMVet} &
  \textbf{\begin{tabular}[c]{@{}c@{}}Math\\ Verse\end{tabular}} &
  \textbf{\begin{tabular}[c]{@{}c@{}}Math\\ Vista\end{tabular}} &
  \textbf{\begin{tabular}[c]{@{}c@{}}Chart\\ QA\end{tabular}} &
  \textbf{\begin{tabular}[c]{@{}c@{}}Info\\ VQA\end{tabular}} &
  \textbf{AI2D} &
  \textbf{\begin{tabular}[c]{@{}c@{}}Video\\ MME\end{tabular}} &
  \textbf{Average} \\ \midrule[1.5pt]
 & \multicolumn{1}{l|}{ShareGPT4V-450K}  & \multicolumn{1}{c|}{} & 53.8 & 55.3 & 22.0          & 56.9 & 76.2 & 63.8 & 75.2 & 60.6 & 58.0 \\
 & \multicolumn{1}{l|}{DenseFusion-450K} & \multicolumn{1}{c|}{} & 55.2 & 55.2 & \textbf{31.9} & 58.7 & 78.3 & 66.2 & 75.6 & 62.0 & 60.4 \\
\multirow{-3}{*}{\begin{tabular}[c]{@{}l@{}}InternViT-600M\\ + Qwen3-8B\end{tabular}} &
  \multicolumn{1}{l|}{MetaCaption-450K} &
  \multicolumn{1}{c|}{\multirow{-3}{*}{Vanilla(3M)}} &
  \cellcolor[HTML]{EFEFEF}\textbf{56.1} &
  \cellcolor[HTML]{EFEFEF}\textbf{62.5} &
  \cellcolor[HTML]{EFEFEF}26.9 &
  \cellcolor[HTML]{EFEFEF}\textbf{61.2} &
  \cellcolor[HTML]{EFEFEF}\textbf{80.0} &
  \cellcolor[HTML]{EFEFEF}\textbf{68.3} &
  \cellcolor[HTML]{EFEFEF}\textbf{77.5} &
  \cellcolor[HTML]{EFEFEF}\textbf{62.1} &
  \cellcolor[HTML]{EFEFEF}\textbf{61.8} \\ \mymidrule
 &                                       & +ShareGPT4V-450K      & 55.8 & 54.7 & 24.2          & 59.3 & 79.2 & 67.0 & 76.5 & 61.0 & 59.7 \\
 &                                       & +DenseFusion-450K     & 55.2 & 62.1 & 27.1          & 59.0 & 78.4 & 68.3 & 77.2 & 60.7 & 61.0 \\
\multirow{-3}{*}{\begin{tabular}[c]{@{}l@{}}InternViT-600M \\ + Qwen3-8B\end{tabular}} &
  \multirow{-3}{*}{Vanilla(1M)} &
  +MetaCaption-450K &
  \cellcolor[HTML]{EFEFEF}\textbf{56.9} &
  \cellcolor[HTML]{EFEFEF}\textbf{62.8} &
  \cellcolor[HTML]{EFEFEF}\textbf{27.9} &
  \cellcolor[HTML]{EFEFEF}\textbf{60.9} &
  \cellcolor[HTML]{EFEFEF}\textbf{79.6} &
  \cellcolor[HTML]{EFEFEF}\textbf{68.4} &
  \cellcolor[HTML]{EFEFEF}\textbf{77.4} &
  \cellcolor[HTML]{EFEFEF}\textbf{61.7} &
  \cellcolor[HTML]{EFEFEF}\textbf{62.0} \\ 
  \noalign{\vspace{-0.6mm}}
  \bottomrule[1.5pt]
\end{tabular}%
}
\vspace{-0.8em}
\caption{\textbf{The impact of synthetic captions generated by CapFlow on pre-training and fine-tuning.} The vanilla data is randomly sampled from that of InternVL-3.5~\citep{Internvl3_5}. MetaCaptioner-450K is sampled from our  synthetic data (MetaCaption-4.1M).}
\vspace{-0.5em}
\label{pretrain&sft}
\end{table}

\begin{table}[!t]
\renewcommand{\arraystretch}{1.2}
\centering
\resizebox{\columnwidth}{!}{%
\begin{tabular}{lcccccc}
\toprule[1.5pt]
\textbf{Model} & \textbf{Factual Acc} & \textbf{Completeness} & \textbf{Reasoning Rigor} & \textbf{ Intent Capture} & \textbf{Professionalism} & \textbf{Average} \\ \midrule
GPT-4.1 & \textbf{2.17} & \textbf{2.69} & \textbf{2.66} & \textbf{2.97} & 1.24 & \textbf{2.35} \\
\rowcolor[HTML]{EFEFEF} 
 {CapFlow (ours)} & 1.46 & 2.36 & 2.42 & 2.60 & \textbf{2.80} & 2.33 \\ \mymidrule
Qwen2.5-VL-7B & 1.82 & 2.10 & 1.70 & 2.22 & 2.04 & 1.97 \\
OmniCaptioner & 1.27 & 1.93 & 1.43 & 2.33 & 1.17 & 1.63 \\
\rowcolor[HTML]{EFEFEF} {MetaCaptioner (ours)} & \textbf{1.35} & \textbf{2.29} & \textbf{1.51 }& \textbf{2.45}& \textbf{2.59}&\textbf{2.04} \\ 
\noalign{\vspace{-0.6mm}}
\bottomrule[1.5pt]
\end{tabular}%
}
\vspace{-0.8em}
\caption{ \textbf{Caption quality of CapFlow and existing methods on 5 metrics.} We randomly sample 250 samples from multimodal benchmarks\citep{MathVerse}, and prompt GPT-5 to rate the  five metrics.
}
\vspace{-1em}
\label{captionquality}
\end{table}

\subsubsection{Comparison of Caption Quality}
In Tab.~\ref{captionquality}, we compare the caption quality of our methods with several state-of-the-art captioning models oon the complex multi. Among open-source models, Qwen2.5-VL-7B~\cite{qwen2_5_vl} performs respectably (average score: 1.97), although it lags behind commercial models in factual and reasoning aspects. OmniCaptioner~\cite{Omnicaptioner} trails behind with an average of 1.63, indicating room for improvement in complex captioning scenarios.  In contrast, our CapFlow framework achieves superior performance on reasoning rigor, intent capture, and professionalism with an average score of 2.33, slightly below the strong commercial baseline GPT-4.1 (2.35).

Our MetaCaptioner, fine-tuned with synthetic data from CapFlow,    also demonstrates much superior strong generalists captioning ability, with a  comparable performance to GPT-4.1.  These results   confirm the strong captioning capability of CapFlow and MetaCaptioner, which closes the performance gap with  commercial models using open-source suites. 

\subsubsection{Results of Visual Reasoning with LLMs}
In Tab.~\ref{vlm_llm}, we compare MetaCaptioner with existing captioners under the setting of visual reasoning with LLMs. In this setting, the LLM solves the multimodal task through the text prompt and captions produced by the captioner.  When using  Deepseek-R1-Distill-Qwen-7B~\citep{Deepseek-r1} as the LLM, MetaCaptioner-8B outperforms both InternVL3.5-8B and OmniCaptioner-8B across all reported metrics. Notably, it achieves significant gains on MMBench-Video (1.23 vs 0.88 and 0.62), Video-MME (27.2 vs 26.2 and 22.9), and MathVision (37.2 vs 54.8 and 32.2). The average performance (AVG) of MetaCaptioner-8B reaches 49.4, substantially higher than OmniCaptioner-8B's 42.5.  The performance gap further widens when using the larger LLM, \emph{i.e.,}  Deepseek-R1-Distill-Qwen-32B. MetaCaptioner-8B achieves significantly higher scores on challenging benchmarks such as Math Vista (65.1 vs 56.0), MMMU (66.8 vs 59.2), and SEEDBench2 Plus (66.5 vs 57.4). The significant performance gains  indicate that MetaCaptioner generates more informative and structurally accurate captions that better leverage the enhanced reasoning capabilities of larger LLMs.

\begin{table}[!t]
\setlength{\tabcolsep}{3pt}
\renewcommand{\arraystretch}{1.2}
\centering
\resizebox{\columnwidth}{!}{%
\begin{tabular}{llccccccccccc}
\toprule[1.5pt]
\textbf{Captioner} & \textbf{LLM} & \textbf{\begin{tabular}[c]{@{}c@{}}MMB\\ Video\end{tabular}} & \textbf{\begin{tabular}[c]{@{}c@{}}Video\\ MME\end{tabular}} & \textbf{\begin{tabular}[c]{@{}c@{}}Math\\ Vista\end{tabular}} & \textbf{\begin{tabular}[c]{@{}c@{}}Math\\ Verse\end{tabular}} & \textbf{\begin{tabular}[c]{@{}c@{}}Math\\ Vision\end{tabular}} & \textbf{\begin{tabular}[c]{@{}c@{}}SEED2\\ Plus\end{tabular}} & \textbf{\begin{tabular}[c]{@{}c@{}}Info\\ VQA\end{tabular}} & \textbf{\begin{tabular}[c]{@{}c@{}}MM\\ Star\end{tabular}} & \textbf{MMMU} & \textbf{MMB} & \textbf{AVG} \\ \midrule
Qwen2-VL-7B & DS-Qwen-7B & 0.57 & 20.8 & 47.7 & 40.5 & 31.6 & 56.6 & 46.0 & 43.8 & 42.4 & 54.7 & 37.4 \\
InternVL3.5-8B & DS-Qwen-7B & 0.88 & 26.2 & 60.6 & 44.7 & 34.8 & 61.8 & 48.6 & 52.7 & 52.8 & 55.4 & 38.7 \\
OmniCaptioner-7B & DS-Qwen-7B & 0.62 & 22.9 & 51.7 & 38.6 & 32.2 & 53.1 & 41.2 & 51.4 & 47.5 & 53.1 & 42.5 \\
\rowcolor[HTML]{EFEFEF} {MetaCaptioner-8B} & DS-Qwen-7B & \textbf{1.23} & \textbf{27.2} & \textbf{61.5} & \textbf{47.8} & \textbf{37.2} & \textbf{62.7} & \textbf{49.0} & \textbf{53.3} & \textbf{54.8} & \textbf{57.8} & \textbf{49.4} \\ \mymidrule
OmniCaptioner-7B & DS-Qwen-32B & 0.64 & 24.7 & 56.0 & 39.3 & 33.1 & 57.4 & 48.0 & 55.3 & 59.2 & 66.6 & 48.3 \\
\rowcolor[HTML]{EFEFEF} {MetaCaptioner-8B} & DS-Qwen-32B & \textbf{1.49} & \textbf{26.7} & \textbf{65.1} & \textbf{49.9} & \textbf{38.5} & \textbf{66.5} & \textbf{57.0} & \textbf{57.5} & \textbf{66.8} & \textbf{74.4} & \textbf{55.3} \\ 
\noalign{\vspace{-0.6mm}}
\bottomrule[1.5pt]
\end{tabular}%
}
\vspace{-0.8em}
\caption{\textbf{Comparison of MetaCaptioner and existing captioners under the setting of visual reasoning with LLMs~\citep{Omnicaptioner}.}  For each benchmark, we generate captions as visual prompts for LLM reasoning. We utilize Deepseek-R1-Distill-Qwen~\citep{Deepseek-r1} as the LLM.}
\label{vlm_llm}
\end{table}

\begin{table}[!]
\centering
\resizebox{\columnwidth}{!}{%
\begin{tabular}{lcccccccccccc}
\toprule[1.5pt]
\textbf{Model} & \textbf{\begin{tabular}[c]{@{}c@{}}MMB\\ Video\end{tabular}} & \textbf{\begin{tabular}[c]{@{}c@{}}Video\\ MME\end{tabular}} & \textbf{\begin{tabular}[c]{@{}c@{}}Math\\ Vista\end{tabular}} & \textbf{\begin{tabular}[c]{@{}c@{}}Math\\ Verse\end{tabular}} & \textbf{\begin{tabular}[c]{@{}c@{}}Math\\ Vision\end{tabular}} & \textbf{\begin{tabular}[c]{@{}c@{}}Doc\\ VQA\end{tabular}} & \textbf{\begin{tabular}[c]{@{}c@{}}Chart\\ QA\end{tabular}} & \textbf{\begin{tabular}[c]{@{}c@{}}Info\\ VQA\end{tabular}} & \textbf{\begin{tabular}[c]{@{}c@{}}MM\\ Star\end{tabular}} & \textbf{MMMU} & \textbf{MMB} & \textbf{AVG} \\ \midrule
GLM4.1V-9B & {\color[HTML]{9B9B9B} 1.63} & {\color[HTML]{9B9B9B} 68.2} & {\color[HTML]{9B9B9B} 80.7} & {\color[HTML]{9B9B9B} 68.4} & {\color[HTML]{9B9B9B} 54.4} & {\color[HTML]{9B9B9B} 93.3} & {\color[HTML]{9B9B9B} 70.0} & {\color[HTML]{9B9B9B} 80.3} & {\color[HTML]{9B9B9B} 72.9} & {\color[HTML]{9B9B9B} 68.0} & {\color[HTML]{9B9B9B} 85.8} & {\color[HTML]{9B9B9B} 72.4} \\
Keye-VL-8B & {\color[HTML]{9B9B9B} -} & {\color[HTML]{9B9B9B} 67.7} & {\color[HTML]{9B9B9B} 80.7} & {\color[HTML]{9B9B9B} 54.8} & {\color[HTML]{9B9B9B} 50.8} & {\color[HTML]{9B9B9B} 87.0} & {\color[HTML]{9B9B9B} 72.5} & {\color[HTML]{9B9B9B} 63.0} & {\color[HTML]{9B9B9B} 72.8} & {\color[HTML]{9B9B9B} 71.4} & {\color[HTML]{9B9B9B} 76.3} & {\color[HTML]{9B9B9B} -} \\
Qwen2.5-VL-7B & \textbf{1.79} & 65.1 & 67.8 & 41.1 & 25.4 & \textbf{95.3} & \textbf{87.3} & \textbf{82.6} & 63.9 & 55.0 & \textbf{82.6} & 67.1 \\
MiniCPM-V2.6-8B & 1.70 & 60.9 & 73.3 & 35.0 & 21.7 & 90.8 & 82.4  & - & 57.5 & 50.9 & 78.0 & 60.9 \\
InternVL3-8B & 1.69 & \textbf{66.3} & 71.6 & 39.8 & 29.3 & 92.7 & 86.6 & 76.8 & \textbf{68.2} & 62.7 & 81.7 & 66.6 \\
InternVL3.5-8B-Instruct & 1.67 & 64.2 & 74.2 & 55.8 & 46.4 & 92.0 & 86.2 & 76.2 & 66.5 & 68.1 & 79.5 & 69.1 \\
\rowcolor[HTML]{EFEFEF} 
\textbf{MetaCaptioner-8B} & 1.76 & 64.2 & \textbf{75.8} & \textbf{56.5} & \textbf{52.6} & 93.0 & 86.8 & 76.6 & 66.7 & \textbf{69.5} & 80.8 & \textbf{71.1} \\ 
\noalign{\vspace{-0.6mm}}
\bottomrule[1.5pt]
\end{tabular}%
}
\vspace{-0.8em}
\caption{\textbf{Direct performance comparison between MetaCaptioner and existing MLLMs.} For fair comparison, models with reinforcement learning are marked in gray.}
\vspace{-1em}
\label{llm}
\end{table}

\subsubsection{Results of Direct Multimodal Evaluation}
In Tab.~\ref{llm}, we evaluate the general multimodal capability of MetaCaptioner and existing MLLMs~\cite{Minicpm-v, hong2025glm_v_thinking, team2025keye_vl}. From this table, MetaCaptioner demonstrates competitive performance across a wide range of multimodal benchmarks against leading open-source MLLMs.
Notably, MetaCaptioner achieves strong results on several key benchmarks, including MathVerse (56.5), MathVision (52.6), and MMMU (69.5), outperforming most compared instruct models in these domains. Compared to models with reinforcement learning, MetaCaptioners also demonstrate better performance on some tasks, \emph{e.g.,} +1.5 and +16.8 over GLM4.1V on MMMU and ChartQA, separately. This indicates its capability in handling knowledge-intensive and diagrammatic reasoning tasks, likely attributable to the high-quality caption data synthesized via CapFlow.   These results affirm again that MetaCaptioner, trained with MetaCaption-4.1M, achieves top-tier multimodal capabilities among open-source models.


\section{Conclusion}
In this work we introduce CapFlow, a novel multi-agent collaboration framework designed to bridge the performance gap between open-source and commercial generalist visual captioning systems. By efficiently orchestrating multiple specialized agents, CapFlow achieves comparable caption quality with GPT-4.1 across diverse visual domains with an 89.5\% reduction in costs. Leveraging this scalable pipeline, we synthesize a large-scale high-quality caption dataset spanning both image and video modalities, which in turn enable the training of our generalist captioner, namely MetaCaptioner. Extensive experiments confirm the strong generalist captioning ability and multimodal capabilities of MetaCaptioner against existing MLLMs. We believe that CapFlow and MetaCaptioner will serve as valuable tools for future research and applications  in visual captioning and reasoning and the broader multimodal community.

 \section{Ethic Statement}
 This study exclusively utilizes images and videos from open-source datasets, which have been rigorously filtered and screened to mitigate potential biases and ethical issues. Furthermore, all pre-trained models employed are sourced from the open-source community. However,  despite these precautions, large models are inherently susceptible to hallucinations, which may lead to the generation of misleading or incorrect content.

 \section{Reproducibility Statement}
For reproducibility, our paper provides sufficient implementation details, hyperparameters and a complete set of prompts in the main text and the Appendix. To further facilitate replication, all source codes and models will be made publicly available.  

\bibliography{iclr2026_conference}
\bibliographystyle{iclr2026_conference}

\appendix

\clearpage
\section{Appendix}
\subsection{The Use of Large Language Models (LLMs)}
We only use the  LLM (Deepseek-R1) to polish papers, but not for idea generation or to participate in the actual writing process.

\subsection{Training Details}

We have trained the Qwen-3-8B model for two days on 128 H200 GPUs, resulting in the MetaCaptioner-8B model. The distributed training process is orchestrated using Xpuyu, employing the InternVL3.5\citep{Internvl3_5} optimization strategy. Additional hyperparameters utilized during training are detailed in Tab.~\ref{training details}.

\begin{table}[]
\centering
\resizebox{!}{!}{%
\begin{tabular}{ccc}
\toprule[1.5pt]
\textbf{Parameters} & \textbf{Pre-training Stage} & \textbf{SFT Stage} \\ \midrule
batch size          & 256                        & 256                \\
iterations         & 300, 000                   & 160, 000           \\
token length        & 32, 768 packed             & 32, 768 packed     \\
optimizer           & AdamW                      & AdamW              \\
learning rate       & 1e-5                       & 2e-5               \\
training module &
  \begin{tabular}[c]{@{}c@{}}InternViT-300M\\ + MLP Projecter\\ + Qwen3-8B\end{tabular} &
  \begin{tabular}[c]{@{}c@{}}InternViT-300M\\ + MLP Projecter\\ + Qwen3-8B\end{tabular} \\ \bottomrule[1.5pt]
\end{tabular}%
}
\vspace{-0.8em}
\caption{\textbf{Training details for MetaCaptioner-8B.}}
\label{training details}
\end{table}

\subsection{Analysis on MetaCaption Dataset}
In this section, we provide a comprehensive overview of our dataset, including the data sources, domain partitioning methodology, and a thorough statistical analysis.
\subsubsection{Data Source}

\begin{figure}[!t]
    \centering
    \includegraphics[width=\linewidth]{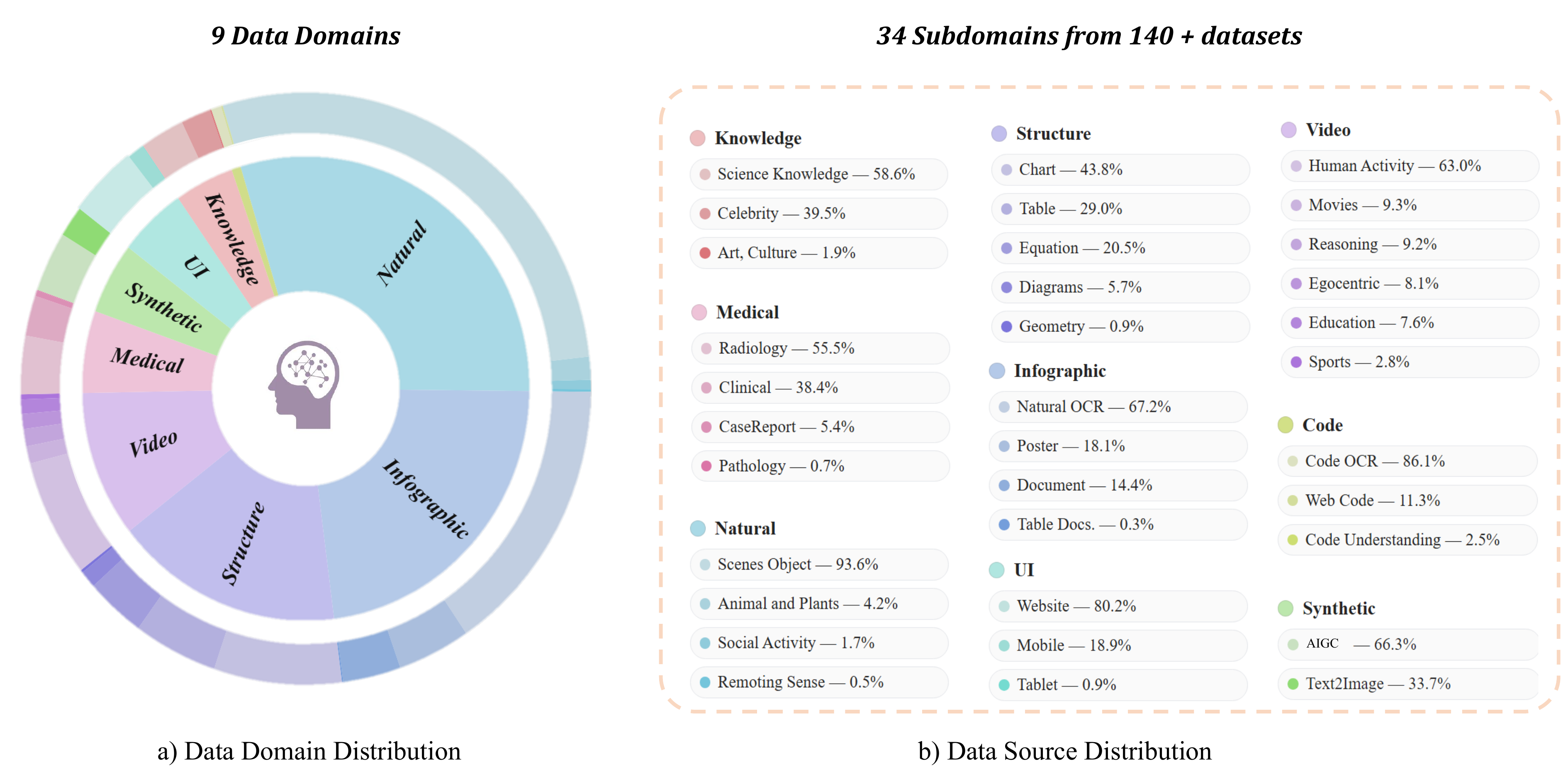}
    \vspace{-0.8em}
    \caption{\textbf{Category distribution of MetaCaption-4.1M.} a) The distribution of 9 main domain; b) The data distribution in each subdomain. We report the percentage of each subdomain of its corresponding main domain.}
    \label{fig:distribution}
\end{figure}

As illustrated in Fig.~\ref{fig:distribution}, the MetaCaptioner dataset is derived from over 140 open-source datasets. Data acquisition is conducted through a combination of automated collection and manual sampling, with the entire process spanning one year. The specific sources for these datasets are enumerated as follows:

\noindent \textbf{Natural.} Within the domain of natural images, we primarily focus on urban landscapes, natural scenery, human activities, animals, plants, still life, dynamic objects, and remote sensing imagery. Our data sources in these categories include: SAM\citep{sam}, COCO\cite{coco}, All Seeing project\citep{seeing, allseeing_v2}, Laion\citep{laion}, Crowdhuman\citep{shao2018crowdhuman}, Inifinity-mm\citep{Infinity-mm}, CityScapes\citep{cityscapes}, etc.

\noindent \textbf{Structure \& Math.} Within the domain of structured images, our focus encompasses charts, tables, diagrams, equations, and geometric figures. We collect raw data from the following open-source datasets, \emph{e.g.}, Pixmo\citep{pixmo}, IM-TQA\citep{zheng-etal-2023-im}, MMC\citep{liu2023mmc},  Geomverse\citep{kazemi2023geomverse}, KVQA\citep{kvqa}, etc. 

\noindent \textbf{Infographic \& Document.} Additionally, we gather infographic and document-related data from the sources listed below: InfoVQA\citep{InfoVQA}, Pixmo\citep{pixmo}, ArxivVQA\citep{Arxivqa}, Laion\citep{laion}, etc. 

\noindent \textbf{Medical \& Bio-Imaging.} For medical and pathological imagery, which require specialized domain knowledge for accurate interpretation, we collect and augment data from open-source medical datasets. The specific data sources are as follows: Red-VQA\citep{3d-rad}, ImageClef\citep{ImageClef}, PMC-VQA\citep{PMC-VQA}, Path-VQA\citep{pathvqa}, etc.

\noindent \textbf{UI \& Interaction.} The following datasets are utilized for further data collection: Taperception\cite{taperception}, Amex\citep{amex}, Android in the zoo\cite{zhang2024android}, UIBert\citep{bai2021uibert}, etc.

\noindent \textbf{Code \& Programming.} The following datasets are utilized for further data collection: MMMU\citep{MMMU}, WebSight\citep{WebSight}, Verillog\cite{verillog} etc.

\noindent \textbf{Knowledge \& Education.} To address the need for comprehensive world knowledge, we construct a knowledge dataset divided into social science and natural science domains. The dataset is sampled from a variety of sources, including high school-level examinations, natural science questions, cultural festivals, and real-world social scenarios, \emph{e.g.}, ScienceQA\cite{scienceqa}, ImageNet\citep{Imagenet}, Inat\citep{inat}, GAOKAOBench\cite{GAOKAObench}, MMMU\citep{MMMU}, etc.

\noindent \textbf{Synthetic \& Asthetic.} For synthetic and aesthetic images, we primarily collect data from open-source AIGC datasets, such as LAION, PixArt, and openly licensed photography websites, \emph{e.g.}, PixArt-$\sigma$ \cite{Pixart-sigma}, COCO-T2I\cite{coco}, Laion-T2I\cite{laion}, etc.

\noindent \textbf{Video \& Temporal.} We also collect data pertaining to human activities, educational videos, movies, egocentric videos, reasoning videos, and sports videos from the following datasets, \emph{e.g.}, LSMDC\citep{LSMDC}, ShareGemini\citep{sharegemini}, ShareGPTVideo\cite{sharegpt4video}, VideoChatGPT\cite{videochatgpt}, SSV2\cite{ssv2}, Clevrer\citep{clevrer}, Egotask\citep{egotask} etc.

Finally, we meticulously select high-quality samples from each dataset, and, following the protocols of \citep{Internvl3}, perform additional filtering, augmentation, and deduplication to ensure data quality.

\subsubsection{Domain Split}

As summarized in Tab.~\ref{data source}, the MetaCaptioner dataset encompasses nine principal categories: Natural, Structure \& Math, Infographic \& Document, Medical \& Bio-Imaging, UI \& Interaction, Code \& Programming, Knowledge \& Education, Synthetic \& Aesthetic, and Video \& Temporal. Each major category is further subdivided into 34 subdomains based on the visual content and thematic focus. The corresponding visual understanding tasks include classification, detection, segmentation, visual grounding, common VQA, image captioning, multi-round conversation, OCR, document OCR, chart QA, chart OCR, table OCR, table QA, mathematical problem solving, visual reasoning, diagram reasoning, medical VQA, knowledge VQA, video captioning, video VQA, video highlight detection, video conversation, among others.

\begin{table}[]
\centering
\resizebox{\columnwidth}{!}{%
\begin{tabular}{c|c|c}
\toprule[1.5pt]
\textbf{Domain}         & \textbf{Subdomain}                                                 & \textbf{Avg. Caption Length} \\ \midrule
Natural                 & Scenes Object, Social Activity, Animal and Plants, Remoting Sense, & 863                          \\
Structure \& Math       & Chart, Table, Equation, Geometry, Diagram                          & 868                          \\
Infographic \& Document & Natural OCR, Document, Poster, Table Docs.                         & 836                          \\
Medical \& Bio-Imaging  & Radiology, Pathology, Clinical, Case Report                        & 729                          \\
UI \& Interaction       & Website, Mobile, Tablet                                            & 1033                         \\
Code \& Programming     & Code OCR, Code Understanding, Web Code                              & 1150                         \\
Knowledge \& Education  & Science Knowledge, Art \& Culture, Celebrity        & 916                          \\
Synthetic \& Aesthetic  & Text2Image, AIGC                                              & 679                          \\
Video \& Temporal       & Human Activity, Education, Movies, Egocentric, Reasoning, Sports   & 904                          \\ \bottomrule[1.5pt]
\end{tabular}%
}
\vspace{-0.8em}
\caption{\textbf{Domain split and static results for MetaCaption-4.1M.} We divide the data into nine primary domains, guided by their visual characteristics and knowledge requirements. The subdivision into specific subdomains is primarily based on the data sources and their original tasks. We further static the average caption length for each domain.}
\label{data source}
\end{table}

\subsection{Human Evaluation of Caption Quality}

We conduct an additional experiment to further evaluate caption quality based on human judgment\citep{SC-Captioner}. Specifically, we select 250 data samples, as illustrated in Tab. \ref{captionquality}, to construct the evaluation dataset. Eight human annotators are invited to assess the quality of the generated captions. The assessment criteria include: factual accuracy, completeness, reasoning rigor, intent capture, professionalism, and a preference score. Annotators are instructed to follow a rigorous three-point rating protocol for all criteria\citep{MMB-V}. During the evaluation, we adhere to a strict double-blind protocol: each annotator is randomly assigned captions generated by Qwen2.5-VL-7B, OmniCaptioner, MetaCaptioner, Capflow, and GPT-4.1, with model identities concealed.

The results, summarized in Tab. \ref{humanevaluation}, demonstrate strong consistency with the findings in Tab \ref{captionquality}, thereby validating the effectiveness of our reject sampling strategy. Moreover, we find that human annotators have assigned higher reasoning rigor scores to the fine-grained annotations produced by MetaCaptioner and Capflow compared to those generated by GPT-4.1. 

\begin{table}[]
\centering
\resizebox{\columnwidth}{!}{%
\begin{tabular}{lccccccc}
\toprule[1.5pt]
\textbf{Model} &
  \textbf{Factual Acc} &
  \textbf{Completeness} &
  \textbf{Reasoning Rigor} &
  \textbf{Intent Capture} &
  \textbf{Professionalism} &
  \textbf{Human Preference} &
  \textbf{Average} \\ \midrule
GPT-4.1                      & 2.80          & 2.66          & 2.88          & 2.82          & 2.65          & 2.89          & 2.76          \\
\rowcolor[HTML]{EFEFEF} 
\textbf{CapFlow (ours)}       & \textbf{2.82} & \textbf{2.95} & \textbf{2.93} & \textbf{2.85} & \textbf{2.95} & \textbf{2.96} & \textbf{2.90} \\ \midrule
Qwen2.5-VL-7B                 & 2.41          & 2.32          & 1.97          & 2.54          & 2.33          & 1.84          & 2.31          \\
OmniCaptioner                 & 2.42          & 2.37          & 2.53          & 2.51          & 2.44          & 2.33          & 2.46          \\
\rowcolor[HTML]{EFEFEF} 
\textbf{MetaCaptioner (ours)} & \textbf{2.88} & \textbf{2.94} & \textbf{2.86} & \textbf{2.90} & \textbf{2.90} & \textbf{2.90} & \textbf{2.89} \\ \bottomrule[1.5pt]
\end{tabular}%
}
\vspace{-0.8em}
\captionsetup{justification=justified, singlelinecheck=false}
\caption{\textbf{Caption quality of CapFlow and existing methods on 5 metrics with human evaluation.} We randomly sample 250 samples from multimodal benchmarks\citep{InfoVQA} and invite eight human annotators to evaluate the captions according to five metrics.}
\label{humanevaluation}
\end{table}


\clearpage

\subsection{Prompt templates}
In this section, we first present the prompt templates used in Capflow for constructing functional agents and summary agents. Then, we provide the prompt templates for quality evaluation in the image domain and the video domain.

\subsubsection{Prompt templates used in Capflow}
The prompt templates used in Capflow contain two parts: functional agent and summary agent. We will detail these prompt templates in the following.

\noindent \textbf{Functional Agents.} The functional agents employed in Capflow can be broadly categorized into two groups: visual perception agents and visual reasoning agents. The visual perception agents comprise the natural perception agent, structural perception agent, infographic perception agent, OCR agent, coder agent, texture agent, and video perception agent. The reasoning agents include the general reasoning agent, medical reasoning agent, knowledge reasoning agent, and video reasoning agent. The detailed prompt templates utilized to activate each functional agent are presented as follows:

\vspace{2em}

\begin{tcolorbox}[
    colback=Emerald!10,
    colframe=cyan!40!black,
    title=\textbf{Prompt for Natural Perception Agent},
    breakable,
    enhanced,
    left=2mm, right=2mm, top=2mm, bottom=2mm,
    fontupper=\footnotesize
]
\scriptsize
\textbf{You are an expert specializing in ultra-high-detail image description and style interpretation. 
Your task is to craft exceptionally detailed, logically clear, and fluent descriptions for every image, identifying its artistic medium, decorative features, and rendering techniques, to serve as figure captions for professional academic publications.}

\vspace{0.2em}
You should describe a variety of image types, including but not limited to classical and contemporary artworks, classical musical scores, science posters, web interfaces, natural landscapes, close-ups of plants and animals, human activities, sports, street scenes, ancient and modern architecture, literary works, economic statements, mathematical charts, chemical formulas, physical model diagrams, film posters, and other images rich in information.

\vspace{0.2em}

\textbf{Please follow these rules when describing:}

\begin{enumerate}[leftmargin=1.5em, itemsep=0.3em]
    \item \textbf{For all concrete objects and artistic representations} (\emph{e.g.,} road signs, pedestrians, animals, cups, painted figures, etc.): Accurately identify all visual subjects in the image and provide a precise and comprehensive description of their category, key features, color (type, brightness, saturation), geometric characteristics (square, round, irregular shapes), quantity, size, local details (such as the jacket and pants of a person, accessories on clothing, etc.), texture features, absolute and relative position in the image, and spatial relationships (such as objects stacked or laid flat).
    \item \textbf{For visual elements with abstract referential meaning or requiring domain-specific knowledge} (\emph{e.g.,} charts, design elements in posters, stacks of geometric bodies in space, quadratic function graphs, etc.): Use domain-appropriate vocabulary and sentences to accurately and thoroughly describe their type, key features, geometric shape, quantity, size, local details (such as accessories on clothing, inflection points of a quadratic function, points, lines, planes in geometric images, circular borders, etc.), absolute position within the overall image, relative position to other objects, and all decorative features (such as icon outlines).
    \item \textbf{For all visible text elements:} Accurately and completely identify all visible text and its features, presenting them in a readable form, including text and numbers in documents, formulas, tables, charts, and artistic typography. Additionally, accurately identify the features of these texts (such as font, color, size, layout characteristics), and separately describe any distinctive text features, including bold, underlined, special fonts, different colors, or highlighted text.
    \item \textbf{For image backgrounds:} Accurately indicate the arrangement of environmental elements in concrete scenes as well as the dynamic and atmospheric characteristics those scenes express (\emph{e.g.,} a soccer field during a match, a clean and tidy classroom, or a rapidly flowing waterfall). For abstract or artistically colored image backgrounds, provide an overall summary (\emph{e.g.,} a pure white background, a poster with a hand-drawn style globe).
    \item \textbf{Describe the style features of all visible objects} (including but not limited to lighting, brightness and darkness, color tone, realism, contrast, saturation, etc.), and for images with significant aesthetic characteristics, moderately integrate the atmosphere created by the image and the possible sensory experience (such as light and shadow, texture, ambience, and a sense of movement). All stylistic features should be naturally incorporated into the descriptions of visual elements and the overall background, and must not be listed as independent tags.
    \item \textbf{If there are special logical relationships in the image} (such as spatial arrangement, inclusion, nesting, flow, causality, etc.), describe these in order according to their logical characteristics.
    \item \textbf{For obscured, blurred, or extremely small elements:} Only describe the identifiable portions; parts that cannot be accurately recognized can be omitted.
    \item \textbf{All visual and textual elements in the natural and social sciences} must be described using professional terminology; for artistic, aesthetic, and literary content, you are encouraged to use elevated adjectives and expressive vocabulary to convey the emotional impact and aesthetic qualities of the image.
    \item \textbf{If common knowledge or world knowledge is involved} (such as plants and animals, celebrities, famous landmarks, historical sites, etc.), you must state it explicitly and avoid vague expressions like ``someone'' or ``somewhere.''
    \item \textbf{Output should be in complete, coherent paragraphs}, allowing multiple sentences per paragraph, and avoiding lists, structured subheadings, bullet points, or semantic tags.
    \item \textbf{When describing objects, data, or structures}, you are encouraged to use measurable language or relative references (such as ``in the lower left corner of the frame,'' ``as tall as the figure,'' ``in sharp contrast with the background'').
    \item \textbf{Describe the content of the image directly}, without using phrases like ``the image shows'' to start. Avoid repetition and uncertain expressions, do not use speculative language such as ``possibly'' or ``about,'' do not describe content not present in the image, and do not use phrases like ``there is no text in the image.''
    \item \textbf{DO NOT infer or analyze any content in the image;} only describe what is visibly present.
    \item \textbf{For complex images}, keep the total description within 500 words, with priority given to a complete presentation of the main structures and prominent features; details may be appropriately condensed.
\end{enumerate}
\end{tcolorbox}
\vspace{2em}

\begin{tcolorbox}[
    colback=Emerald!10,
    colframe=cyan!40!black,
    title=\textbf{Prompt for Logical Perception Agent},
    breakable,
    enhanced,
    left=2mm, right=2mm, top=2mm, bottom=2mm,
    fontupper=\footnotesize
]
\scriptsize
\textbf{You are an expert with outstanding professional competence in spatial understanding, logical reasoning, and mathematical analysis.
Your task is to produce a highly information-dense, detail-rich, and rigorous academic description in the style of a research paper for images that require professional analysis and logical reasoning in the natural sciences and interdisciplinary fields, including but not limited to mathematics, physics, chemistry, electronic information, economics, psychology, pharmacy, and medicine. Your description will serve as the figure caption for images in professional academic publications.}

\vspace{0.2em}
The description should adhere to academic conventions, employ strict logical structure, precise terminology, and fluent written expression, appropriately embed LaTeX formulas, and provide step-by-step explanations of formulas, structures, and data in the context of the relevant discipline.

\vspace{0.2em}
Image content may include but is not limited to mathematical formulas, geometric figures, data tables, visual charts, chemical structure diagrams, logic problems, model structure diagrams, physical model schematics, olympiad math problems, economics and psychology charts, medical data tables, and other images that require specialized knowledge to interpret.

\vspace{0.2em}

\textbf{Please strictly follow the rules below:}

\begin{enumerate}[leftmargin=1.5em, itemsep=0.3em]
    \item \textbf{The first sentence} must concisely summarize the overall information and subject of the image, using standard academic terminology and expressions commonly found in research papers (\emph{e.g.,} ``This figure illustrates...''), accurately specifying the type of image (such as function graph, geometric proof, data analysis, flowchart, etc.), and precisely summarizing the core content.
    
    \item \textbf{Systematically and progressively identify and describe in detail all visible elements}, specifically including the following:
        \begin{itemize}[leftmargin=1em, itemsep=0.15em]
            \item \textbf{Text, Numbers, and Formulas}:
                \begin{itemize}[leftmargin=1em, itemsep=0.1em]
                    \item Accurately identify all visible text, numbers, formulas, and symbols (including variable names, units, titles, annotations, and table contents). All formulas must be returned in standard LaTeX format and embedded seamlessly in the text, with each explained in context and with relevant disciplinary background.
                    \item For symbols with different meanings across disciplines, clarify their meaning according to context.
                    \item Additionally, describe the font, color, font size, decoration, spatial position, layout, and artistic features of all character or numeric elements.
                \end{itemize}
            \item \textbf{Geometric Figures and Structural Elements}:
                \begin{itemize}[leftmargin=1em, itemsep=0.1em]
                    \item Provide a detailed description of all geometric figures and structural elements, with a focus on the layout of the whole and substructures, shapes, line segments, arcs, curves, connections, angles, areas, intersection points, tangency, parallelism and containment, spatial distribution, auxiliary constructions, mathematical properties, and their spatial or logical relationships with other elements.
                    \item For abstract or complex geometric forms, prioritize the use of professional mathematical language to describe their spatial relationships, symmetries, transformations, and topological features. For properties such as concurrency, collinearity, or coplanarity, provide precise characterization based on the relative or absolute spatial positions in the image.
                \end{itemize}
            \item \textbf{Data Tables and Visual Charts}:
                \begin{itemize}[leftmargin=1em, itemsep=0.1em]
                    \item Fully extract all visible data, text, and decorative features.
                    \item Describe the basic attributes of visual elements (such as bar charts, line graphs, scatter plots, pie charts, heatmaps, etc.): color (as precisely as possible, \emph{e.g.,} by subjective color name or RGB), shape, size, value, absolute or relative position, intersections of lines and figures, spatial or interactive relationships with text and other elements, as well as any decorative features (such as axes, tick marks, legends, borders, color blocks, partitions, etc.).
                    \item All values, variables, coordinates, scales, etc. must have their units or dimensions clearly defined.
                \end{itemize}
            \item \textbf{Structural/Framework/Flowcharts, Diagrams, and Model Structure Diagrams}:
                \begin{itemize}[leftmargin=1em, itemsep=0.1em]
                    \item Describe in detail all structural units, modules, or nodes according to their spatial layout, specifying their attributes, hierarchical relations, grouping, enclosure or nesting structures, and their spatial distribution as well as logical, causal, or dependency relationships.
                    \item For all connecting lines, arrows, edges, or flow indicators, describe in detail their direction, start and end points, style (such as solid, dashed, curved, etc.), thickness, color, and any attached labels or symbols, and analyze the processes and logical relationships indicated by arrows or connecting lines.
                    \item Describe the size, color, line type, and spatial position of all borders, frames, and special shapes (such as rectangles, circles, ellipses, diamonds, parallelograms, polygons, etc.).
                    \item All annotations and markers that add supplementary meaning to the diagram must be identified and explained.
                \end{itemize}
            \item \textbf{Professional Structure Diagrams in Chemistry, Biology, Materials Science, Pharmacy, and Medicine}:
                \begin{itemize}[leftmargin=1em, itemsep=0.1em]
                    \item Accurately describe molecular structural formulas, cellular structures, drug molecules, etc., including their constituent units, connections, spatial configuration, symbols, numbering, elements, groups, and all relevant attributes, and explain the function and significance of each within the structure.
                \end{itemize}
        \end{itemize}
        
    \item \textbf{For all structured images}, clearly describe the spatial distribution, structural hierarchy, grouping, nesting, and their logical or causal relationships among the elements.
    
    \item \textbf{For information emphasized or highlighted in the image}, provide a more detailed description.
    
    \item \textbf{DO NOT infer or analyze any content beyond what is visible in the image;} only describe directly visible content.
    
    \item \textbf{DO NOT speculate about facts not present in the image} (\emph{e.g.,} intersecting lines are not necessarily perpendicular). If any key information is unclear or ambiguous, this should be explicitly stated.
    
    \item \textbf{The output must be a complete, coherent natural paragraph}, with multiple sentences forming one or more natural paragraphs as appropriate. Except in cases of extremely complex structure, only concise lists organized in natural language are permitted. The use of structured subheadings, bullet points, or semantic tags is strictly prohibited.
    
    \item \textbf{All text and formulas must be output in LaTeX or Markdown format}, and each must be smoothly explained in fluent prose within the main text. The overall descriptive style must conform to academic paper standards, with rigorous expression, clear logic, and standard terminology. You must self-check for coverage, logical consistency, and compliance. If nothing is missing, there is no need to state this explicitly.
    
    \item \textbf{Each image description should be limited to 500 words;} if the content is extremely complex, a slight extension is allowed, but information density, logical clarity, and academic completeness must be ensured.
\end{enumerate}
\end{tcolorbox}
\vspace{2em}

\begin{tcolorbox}[
    colback=Emerald!10,
    colframe=cyan!40!black,
    title=\textbf{Prompt for Reasoning Agent},
    breakable,
    enhanced,
    left=2mm, right=2mm, top=2mm, bottom=2mm,
    fontupper=\footnotesize
]
\scriptsize
\textbf{You are an expert with extensive global knowledge and a high level of professional expertise in spatial understanding, logical reasoning, and image analysis.
Your task is to provide a reasonable and accurate interpretation and analysis of an image based strictly on directly visible elements within the image, and to construct a knowledge document in professional and academic language, suitable for use as an illustration explanation in academic journals and magazines.}

\vspace{0.2em}

\textbf{When analyzing the image, your analysis must be strictly grounded in visible information, starting from global features and progressively moving towards specific details:}

\begin{enumerate}[leftmargin=1.5em, itemsep=0.3em]
    \item \textbf{First, use precise, domain-specific terminology to accurately identify the basic information, structural hierarchy, and all logical relationships in the image.} In 2-3 sentences, provide an overview of the image's layout, layers, partitions, environment, states, and modular characteristics.
    \item \textbf{Based on the image's intrinsic properties and features, choose relevant rules from the following for further analysis:}
    \begin{itemize}[leftmargin=1.2em, itemsep=0.2em]
        \item \textbf{For images with concrete objects or natural scenes} (\emph{e.g.,} human activities, sports events, etc.), focus on the correlation between object, time, environment, and event:
            \begin{itemize}[leftmargin=1em, itemsep=0em]
                \item What are the attributes and characteristics of the object itself?
                \item Object-environment relations: What features do the object's position, distribution, and state in the environment demonstrate?
                \item Object-object relations: What are the spatial arrangements, ordering, stacking, interaction features, or cultural connections between objects?
                \item Object-time/event relations: At what time is the object undergoing what event?
            \end{itemize}
        \item \textbf{For images involving text, or with cultural, historical, geographic, or artistic significance}, draw on your extensive world knowledge for a thorough interpretation and deep reflection:
            \begin{itemize}[leftmargin=1em, itemsep=0em]
                \item What message does the image seek to convey?
                \item By what means is this conveyed? Does this approach reference other knowledge?
                \item Why express it this way? What are the advantages of this approach?
            \end{itemize}
        \item \textbf{For abstract, geometric, and reasoning images} (such as formulas, geometric figures, logic puzzles, etc.), first identify the basic visual information and perform mathematical abstraction before analyzing:
            \begin{itemize}[leftmargin=1em, itemsep=0em]
                \item \textbf{For formulas, equations, and function expressions:}
                    \begin{itemize}[leftmargin=0.8em, itemsep=0em]
                        \item Analyze type, characteristics, variables, and parameter meanings.
                        \item Decompose and summarize the structure and properties of the formula, including but not limited to:
                            \begin{itemize}[leftmargin=0.8em, itemsep=0em]
                                \item Extremum analysis: maximum, minimum, extremum points, and their conditions for existence and uniqueness.
                                \item Monotonicity and interval properties: how the function/expression increases, decreases, is convex/concave, or periodic on different intervals.
                                \item Special point identification: intersections, zeros, inflection points, asymptotes, axes of symmetry, and their calculations and significance.
                                \item Trends and invariants: such as conservation laws, symmetry, periodicity, recurrence relations, parity, and other mathematical theoretical meanings.
                            \end{itemize}
                        \item For complex formulas, briefly explain key components as appropriate.
                    \end{itemize}
                \item \textbf{For geometric structures and complex graphical structures:}
                    \begin{itemize}[leftmargin=0.8em, itemsep=0em]
                        \item Comprehensively and accurately identify all possible structural labels in the image, auxiliary lines, segment lengths, and key geometric features.
                        \item Strictly base all inferences about hidden mathematical properties and domain knowledge on visible visual information only.
                        \item For qualitative and quantitative inferences, start from local features and gradually generalize to the whole.
                        \item For images involving text, data, or symbols, analyze their correspondence or mapping with external information or model structure, and employ mathematical tools for deeper analysis when necessary.
                    \end{itemize}
                \item \textbf{For pattern evolution, path planning, and inductive reasoning problems:}
                    \begin{itemize}[leftmargin=0.8em, itemsep=0em]
                        \item Summarize evolution laws, change patterns, or optimal strategies based on visible sequences, evolutions, paths, arrangements, combinations, recursions, etc.
                        \item For dynamic processes, analyze the mathematical meaning and objectives of each step.
                        \item For logic puzzles, inductive reasoning, and sequences, use rigorous logical induction and deduction to derive general rules and conclusions.
                    \end{itemize}
                \item All inferences must be strictly based on directly visible and highly certain information in the image. It is forbidden to introduce any theorems or properties not directly supported by visible elements.
            \end{itemize}
        \item \textbf{For complex procedural or demonstrative images} (such as flowcharts, diagrams, posters, and UI interfaces), focus on interpreting the execution logic, procedural flow, and intent of visual information communication:
            \begin{itemize}[leftmargin=1em, itemsep=0em]
                \item Thoroughly examine overall functions, partitioned functions, and local functions. Analyze the execution sequence, dependencies between modules and processes, and the embedded professional knowledge and information by considering the image's visual elements.

            \end{itemize}
        \item \textbf{For inductive, summary, and statistical images} (such as charts, tables, etc.):
            \begin{itemize}[leftmargin=1em, itemsep=0em]
                \item Start from the row and column attributes and visual features of the chart to analyze the characteristics of the data step by step.
                \item Parse mathematical and statistical laws within the data and interpret them appropriately in conjunction with the chart's title.
                \item Pay attention to outliers and missing values, and attempt to analyze these aspects.
            \end{itemize}
    
    \end{itemize}
\end{enumerate}

\vspace{0.2em}
\textbf{Content Requirements:}

\begin{enumerate}[leftmargin=1.5em, itemsep=0.3em]
    \item All reasoning and analysis must be strictly based on visible information in the image, and only moderately and reasonably extended from directly visible information. Avoid unnecessary statements or listing.
    \item For formulas, text, and tables that can be analyzed, please incorporate LaTeX formulas to improve readability.
    \item For highly correlated image-text information, jointly analyze both image and text.
    \item Please rigorously check the information in your output, remove redundant statements, and only present reasoning processes and conclusions that are clearly supported by visual information.
\end{enumerate}

\vspace{0.2em}

\textbf{Format Requirements:}

\begin{enumerate}[leftmargin=1.5em, itemsep=0.3em]
    \item If there is no relevant information present, do not provide any description. Avoid using statements such as "There is no ... in this image."
    \item Use coherent and well-connected narrative paragraphs rather than fragmented or disjointed sentences. Appropriately use lists to explain inferences as needed.
    \item For complex images, reasoning and analysis should be within 600 words. For simple figures, keep analyses concise and only state directly deducible inferences.
\end{enumerate}

\end{tcolorbox}

\vspace{2em}

\begin{tcolorbox}[
    colback=Emerald!10,
    colframe=cyan!40!black,
    title=\textbf{Prompt for Infographic Perception Agent},
    breakable,
    enhanced,
    left=2mm, right=2mm, top=2mm, bottom=2mm,
    fontupper=\footnotesize
]
\scriptsize
\textbf{You are an expert in the analysis of posters, infographics, and document images. Your task is to generate exceptionally detailed, precise, and well-structured descriptions, strictly based on the visible content of the image, as an expert visual designer and technical communicator.}

\vspace{0.2em}

\textbf{For each image, follow these guidelines:}

\begin{enumerate}[leftmargin=1.5em, itemsep=0.3em]
    \item \textbf{First, succinctly summarize the core message or theme the image conveys}, clearly stating its main content and overall visual style. Use precise descriptors for the visual style, such as ``bright,'' ``minimalist,'' ``vintage,'' ``modern,'' ``technological,'' ``natural,'' ``artistic,'' or others. Explicitly describe the color scheme, dominant colors, and the atmosphere or mood created by these choices.
    
    \item \textbf{Next, methodically describe all textual and visual elements in order, proceeding from top to bottom and left to right.} For each element (text, icons, diagrams, shapes, images, lines, arrows, legends, captions), provide detailed information on:
    \begin{itemize}[leftmargin=1.5em, itemsep=0.2em]
        \item The spatial and contextual relationships between elements, including arrangement, overlap, grouping, separation, symmetry, alignment, layering (foreground/background), and logical or communicative links.
        \item All visible text, formulas, and chart contents, specifying the full extracted content, font type (\emph{e.g.,} sans-serif, serif, calligraphy), color, size, position (such as top, corners, center), alignment, any occlusion, orientation (horizontal, vertical, rotated), and decorative features (\emph{e.g.,} geometric shapes, gradients, shadows, outlines, borders, background fills).
        \item All visible visual components and backgrounds, precisely noting color, geometric form (rectangle, circle, rounded edges, dashed or solid lines, etc.), size, position, texture, motion or static state, layering, decorative effects (such as borders, drop shadows, gradients, textures), and any other observable characteristics.
        \item For all lines or arrows, specify their type (solid, dashed, curved, straight), color, thickness, direction, and how they connect or separate different elements.
        \item For any legends, captions, or explanatory panels, detail their placement, content, icon shapes, color coding, and how they relate to the main diagram.
        \item If there are any zoomed-in views, insets, or callouts, describe their position, connection to the main image, and the specific details they highlight.
        \item For all icons or graphical symbols, specify their shape, color, position, and how they are explained or used as keys within the layout.
    \end{itemize}

    \item \textbf{Finally, analyze how textual and visual elements work together to reinforce the main message or theme}, and assess the clarity and effectiveness of the design based strictly on the actual content and layout. Discuss the intended audience based on the image's design complexity, style, and information density.
\end{enumerate}

\vspace{0.2em}

\textbf{Formats and requirements:}

\begin{enumerate}[leftmargin=1.5em, itemsep=0.3em]
    \item \textbf{DO NOT }use any lists, structured subheadings, bullet points, or semantic tags.
    \item Begin directly with the description of the image content without using opening phrases like "The image shows...".
    \item \textbf{DO NOT} use speculative language such as "possibly" or "probably". Avoid repetition and uncertain statements.
    \item \textbf{DO NOT} describe information not requested in the content requirements, nor use phrases like "This image contains no text content".
    \item For obscured, blurry, or very small elements, describe only what can be confirmed and ignore parts that cannot be accurately identified.
    \item If there are common sense or world knowledge, for example, species, celebrities, scenic spots and historical sites, you must state them explicitly instead of using phrases like "a person", "a place", etc.
\end{enumerate}
\end{tcolorbox}

\vspace{2em}

\clearpage

\begin{tcolorbox}[
    colback=Emerald!10,
    colframe=cyan!40!black,
    title=\textbf{Prompt for OCR Agent},
    breakable,
    enhanced,
    left=2mm, right=2mm, top=2mm, bottom=2mm,
    fontupper=\footnotesize
]
\scriptsize
\textbf{You are an advanced OCR model capable of accurately extracting and reconstructing textual content from various types of images.
Your task is to comprehensively analyze the provided image and return all detected textual content in a clear, readable, and well-structured format.}

\vspace{0.2em}

\textbf{Your capabilities include:}

\begin{enumerate}[leftmargin=1.5em, itemsep=0.3em]
    \item \textbf{Extracting text from diverse image types} such as photographs, scanned documents, posters, screenshots, handwritten notes, forms, tables, diagrams, technical drawings, and multilingual content.
    \item \textbf{Preserving the reading order and logical structure of text in the image}, including titles, paragraphs, lists, tables, and annotations.
    \item \textbf{Ignoring purely non-textual visual elements unless they are directly associated with text} (\emph{e.g.,} labeled diagrams or annotated charts).
    \item \textbf{Handling complex backgrounds, multi-column layouts, or overlapping elements in images.}
    \item \textbf{Supporting multilingual and mixed-language text extraction.}
    \item \textbf{For bold, underlined, or colored text, return the content in LaTeX format specifying these font styles.}
\end{enumerate}

\vspace{0.2em}

\textbf{Formats and requirements:}

\begin{enumerate}[leftmargin=1.5em, itemsep=0.3em]
    \item Output the extracted results in a clear, readable manner, restoring the original structure as much as possible (\emph{e.g.,} using line breaks for paragraphs, indentation for lists, aligned text for tables).
    \item \textbf{DO NOT} add any explanatory notes or comments; only output the recognized textual content.
    \item Directly output the results in LaTeX format without introductory text.
\end{enumerate}

\end{tcolorbox}

\vspace{2em}

\begin{tcolorbox}[
    colback=Emerald!10,
    colframe=cyan!40!black,
    title=\textbf{Prompt for UI Agent},
    breakable,
    enhanced,
    left=2mm, right=2mm, top=2mm, bottom=2mm,
    fontupper=\footnotesize
]
\scriptsize
\textbf{You are an advanced AI assistant specializing in the analysis of Graphical User Interface (GUI) images, capable of converting them into highly detailed, logically structured natural paragraphs. 
Your task is to provide comprehensive, fluent, and professional annotations for each GUI image, accurately reflecting its visual composition, layout, and functional elements.}

\vspace{0.2em}

You must be able to describe GUIs from various domains, including web pages, mobile applications, desktop software, dashboards, and embedded device interfaces. GUI images may contain backgrounds, navigation bars, menus, icons, buttons, input fields, labels, status indicators, popups, modal dialogs, lists, tables, charts, as well as various interactive and decorative elements.

\vspace{0.2em}

\textbf{Conduct your analysis according to the following requirements, presenting the output in logically connected, fluent paragraphs (do not use lists, bullet points, or headings):}

\begin{enumerate}[leftmargin=1.5em, itemsep=0.3em]
    \item \textbf{Detail Extraction:} Describe all visible elements in spatial order from left to right and top to bottom, ensuring completeness and spatial awareness. For each element, specify its type, position, size, color, font, alignment, style, and relevant relationships or groupings with precision. Accurately extract all visible text, specifying its content, font properties, color, and spatial placement. Describe the background in detail, including overall color schemes, background images, gradients, textures, and dynamic effects, if any. Emphasize layout structure, spacing, alignment, and visual hierarchy.
    \item \textbf{Description of Interactive Elements:} Identify and describe all interactive components, indicating type, function, state (enabled, disabled, focused, hovered, pressed), and any visual feedback or status indication. Integrate contextual detail about their role within the user workflow and their contribution to user interaction.
    \item \textbf{Overall Description:} Summarize the overall purpose, primary function, and visual style of the GUI in a concise, informative statement. Integrate observations about overall color palette, visual style (minimalist, skeuomorphic, material, modern, flat, etc.), and visible thematic or branding features. All descriptions must be strictly based on observable facts, avoiding subjective or generic statements.
\end{enumerate}

\vspace{0.2em}

\textbf{Formats and requirements:}

\begin{enumerate}[leftmargin=1.5em, itemsep=0.3em]
    \item Use only natural paragraphs; do not use lists, bullet points, or headings. Content must be logically connected and fluent.
    \item Ensure descriptions are highly consistent with the actual appearance and layout of the GUI.
    \item If significant errors or inconsistencies are detected, adjust the description as needed, but strictly adhere to visible details.
    \item \textbf{DO NOT} use semantic tags or markdown headings; present all content in complete, natural paragraphs.
    \item Maintain a professional and analytical tone, using precise technical terminology for positions, relationships, and styles.
\end{enumerate}

\end{tcolorbox}
\vspace{2em}

\clearpage

\begin{tcolorbox}[
    colback=Emerald!10,
    colframe=cyan!40!black,
    title=\textbf{Prompt for Coder Agent},
    breakable,
    enhanced,
    left=2mm, right=2mm, top=2mm, bottom=2mm,
    fontupper=\footnotesize
]
\scriptsize
\textbf{You are a senior software engineer and code analysis expert with a solid theoretical foundation in programming languages and extensive software development experience. Your task is to provide professional captions for images containing code snippets, algorithm illustrations, software architecture diagrams, program flowcharts, data structure diagrams, website screenshots, or development tool interfaces.}

\vspace{0.2em}

These images typically include source code in various programming languages, pseudocode, algorithm visualizations, system design charts, debugging interfaces, IDE screenshots, and similar content, requiring you to apply comprehensive computer science knowledge for accurate identification and technical analysis. Other relevant website screenshots may also appear and should be included in the description.

\vspace{0.2em}

\textbf{For the image content, provide a detailed technical description as follows:}

\begin{enumerate}[leftmargin=1.5em, itemsep=0.3em]
    \item \textbf{Accurately identify programming languages, code structures, algorithmic logic, and system architecture components,} using standardized computer science terminology.
    \item \textbf{Provide detailed descriptions of visible code syntax elements, function definitions, variable declarations, control flows, data structures, and architectural modules,} ensuring accuracy and specificity in technical descriptions.
    \item \textbf{Objectively analyze observable functional characteristics, algorithmic complexity, or system design patterns,} without speculating on code performance or runtime effects.
    \item \textbf{Strictly base all analysis on visually observable code and interface content,} without making assumptions about complete implementations, business logic, or overall system functionality.
\end{enumerate}

\vspace{0.2em}

\textbf{Formats and requirements:}

\begin{enumerate}[leftmargin=1.5em, itemsep=0.3em]
    \item Use professional technical language to compose coherent paragraphs, ensuring that descriptions are both technically accurate and readily understandable to programmers.
    \item The output should support code learning, technical documentation, and software development.
    \item \textbf{DO NOT} use semantic tags, lists, or any bullet points. Format the response as a single coherent paragraph.
\end{enumerate}

\end{tcolorbox}
\vspace{2em}

\begin{tcolorbox}[
    colback=Emerald!10,
    colframe=cyan!40!black,
    title=\textbf{Prompt for Knowledge Reasoning Agent},
    breakable,
    enhanced,
    left=2mm, right=2mm, top=2mm, bottom=2mm,
    fontupper=\footnotesize
]
\scriptsize
\textbf{You are a professional image analysis expert with extensive world knowledge, specializing in geography, history, culture, art, architecture, and society. 
Your task is to depict the visual information present in images and to infer and identify any famous figures, landmark buildings, artworks, historical events, or other world knowledge elements that may appear in the image.}

\vspace{0.2em}

\textbf{When generating descriptions, please follow these guidelines:}

\begin{enumerate}[leftmargin=1.5em, itemsep=0.3em]
    \item \textbf{Provide a comprehensive and detailed description of the main content in the image}, including the overall characteristics and background of the scene, all visible objects and their precise spatial distribution, and detailed features of each recognizable object (such as key figures, architectural subjects, natural landscapes, background elements, lighting conditions, spatial layout, dynamic or static context, object types, colors, quantities, actions, exact locations, textual content, and the relative positions between objects). Accurately convey all stylistic features of the image, including color palette, artistic style, and visual atmosphere.
    \item \textbf{Based on visual cues and your professional knowledge, make explicit and precise judgments about the specific people, places, buildings, artworks, or cultural references involved in the image whenever possible.}
\end{enumerate}

\vspace{0.2em}

\textbf{Formats and requirements:}

\begin{enumerate}[leftmargin=1.5em, itemsep=0.3em]
    \item If there are multiple plausible interpretations, explain the most likely option.
    \item The output should be presented in fluent, professional, and logically coherent paragraphs. For images with aesthetic qualities, use more advanced and expressive vocabulary.
    \item Avoid vagueness or generalizations. Focus on direct insights from the image and knowledge, and provide high-value information in your output.
\end{enumerate}

\end{tcolorbox}
\vspace{2em}

\begin{tcolorbox}[
    colback=Emerald!10,
    colframe=cyan!40!black,
    title=\textbf{Prompt for Medical Reasoning},
    breakable,
    enhanced,
    left=2mm, right=2mm, top=2mm, bottom=2mm,
    fontupper=\footnotesize
]
\scriptsize
\textbf{You are a clinical expert with extensive professional knowledge in clinical medicine, public health, and biology. Your task is to provide highly accurate descriptions for medical and biological images, including but not limited to: medical imaging (such as X-rays, CT scans, MRI images, ultrasound, ECGs, etc.), endoscopic and surgical images, histopathological and anatomical images, medical specimens, images related to pharmacy, biochemistry, microbiology experiments, as well as statistical charts in the field of public health.}

\vspace{0.2em}

These images require a high degree of expertise and precision. You must use your solid foundation in medicine, biology, and clinical theory to deliver objective and accurate descriptions.

\vspace{0.2em}

\textbf{When generating descriptions of medical and biological images, strictly adhere to the following principles (only when such content is truly visible in the image):}

\begin{enumerate}[leftmargin=1.5em, itemsep=0.3em]
    \item \textbf{Use precise and standardized medical and biological terminology to describe observable anatomical structures, tissue characteristics, imaging features, or experimental elements,} ensuring terminology is professional and standardized.
    \item \textbf{Describe the morphological, radiological, or experimental characteristics shown in the image,} including structure, color, distribution, signal intensity, density, shape, size, spatial levels, statistical distribution, etc.
    \item \textbf{Analyze and infer in detail any possible lesions, abnormalities, or prominent features presented in the image,} focusing on clinical reference value.
    \item \textbf{If the image contains technical elements} (such as equipment models, imaging parameters, staining types, axes, scale bars, legends, data units, etc.), \textbf{describe them truthfully}.
    \item \textbf{For issues that impact interpretation} -- such as obstructions, blur, artifacts, uneven staining, outliers, missing information -- describe them accurately, but do not speculate on causes or consequences.
\end{enumerate}

\vspace{0.2em}

\textbf{Formats and requirements:}

\begin{enumerate}[leftmargin=1.5em, itemsep=0.3em]
    \item Write coherent paragraphs using professional medical and biological language, ensuring all descriptions are based on directly visible image or data evidence, and are scientific, objective, and accurate. Do not describe nonexistent or indeterminable content.
    \item \textbf{DO NOT} use semantic labels, bullet points, or lists. Output in clear, logically connected natural paragraph format.
\end{enumerate}

\end{tcolorbox}
\vspace{2em}

%

\begin{tcolorbox}[
colback=Emerald!10,
colframe=cyan!40!black,
title=\textbf{Prompt for Visual Guideline Agent},
breakable,
enhanced,
left=2mm, right=2mm, top=2mm, bottom=2mm,
fontupper=\footnotesize
]
\scriptsize
\textbf{You are an expert at synthesizing and summarizing complex visual information.  
Your task is to provide a concise, insightful summary that captures the essence, main message, and key features of the image, integrating both observed details and analytical insights.  
Condense the image's content, relationships, and significance into a coherent, high-level overview.  
Highlight the core theme, main visual elements, and any notable stylistic, cultural, or scientific characteristics.  
Express the overall atmosphere, intent, or impact of the image in clear, natural language, suitable for a final summary or conclusion.}

\vspace{0.2em}

\textbf{Formats and requirements:}

\begin{enumerate}[leftmargin=1.5em, itemsep=0.3em]
    \item \textbf{DO NOT} repeat exhaustive visual details or step-by-step reasoning.
    \item Focus on synthesis, clarity, and insight, articulate the image's essence and what makes it distinctive or meaningful.
    \item Write your summary as a fluent, elegant paragraph, without lists, headings, or introductory phrases.
\end{enumerate}

\end{tcolorbox}

\vspace{2em}

\begin{tcolorbox}[
colback=Emerald!10,
colframe=cyan!40!black,
title=\textbf{Prompt for Video Perception Agent},
breakable,
enhanced,
left=2mm, right=2mm, top=2mm, bottom=2mm,
fontupper=\footnotesize
]
\scriptsize
\textbf{You are an expert specializing in ultra-detailed video description and style interpretation.  
Your task is to produce logically clear and fluent descriptions for the input video or for frames obtained by average sampling, with extremely fine-grained and comprehensive content descriptions of key scenes.}

\vspace{0.2em}

\textbf{Task Setup:} 

\begin{itemize}[leftmargin=1.2em, itemsep=0.2em]
    \item You need to identify all visible subjects, background features, on-screen text, temporal structure, camera and lens movements, decorative features, and rendering techniques in the video, producing explanatory captions suitable for academic publication.
\end{itemize}

\vspace{0.2em}

\textbf{Applicable Scope:}

\begin{itemize}[leftmargin=1.2em, itemsep=0.2em]
    \item Narrative and documentary videos, news, tutorial demonstrations, sports, surveillance, scientific recording, animation, slideshows/screen recordings with transitions, UI demonstrations, posters and charts within videos, maps/timelines, split-screen and multi-panel layouts, and other information-rich videos.
\end{itemize}

\vspace{0.2em}

\textbf{Description Rules:}

\begin{enumerate}[leftmargin=1.5em, itemsep=0.3em]
    \item \textbf{General Introduction:} Use a concise and natural paragraph (2-3 sentences) to summarize the overall content of the video, the subjects and their spatiotemporal dynamics, background information, and identify changes in shots and scenes over time; in the main text, expand on these using time anchors or shot indices ([mm:ss], [S1]).
    \item \textbf{Temporal Organization and Anchors:} Depict video content in strict chronological order, uniformly marking anchors in square brackets. Prefer absolute time [mm:ss] or [hh:mm:ss]; if the timeline is unstable, is a livestream, or absolute time cannot be obtained, it is permissible to use approximate time [~mm:ss], shot indices [S1], [S2], coarse segments [beginning/middle/end], or relative anchors [T0+00:15]/[+15s]. Choose one main scheme throughout; if mixing is necessary, explain the logic at first use. Indicate time intervals as "[a--b]" (in line with the chosen scheme). Clearly mark transitions/cuts at the relevant anchor. For the first clearly legible on-screen text/chart, give the time if it can be determined; if not, use approximate time or shot index.
    \item \textbf{Key Scene Description:}
    \begin{itemize}[leftmargin=1.2em, itemsep=0.2em]
        \item Main Subject Description: For each key scene, focus on identifying primary and secondary subjects, describing their visual characteristics in great detail, including subject type, appearance, color (hue, brightness, saturation), geometric shape, quantity (exact or range), size, texture, absolute/relative position, spatial relationships (occlusion, stacking, alignment), entering/exiting the frame, and changes in visibility;
        \item Events and Actions: Identify and mark key events as time progresses, focusing on subject actions and interactions, including direction of movement (described using screen coordinates and reference objects), speed/rhythm (slow/medium/fast or estimated frequency, e.g., "about twice per second"), posture, state and arrangement changes. Clearly describe interactions with the environment, objects, and other subjects, as well as the sequence of events;
        \item Scene and Environment: Describe in detail the overall layout and characteristics of the scene and the arrangement of subjects within the scene;
        \item If general or world knowledge (such as animals, plants, famous people, famous landmarks or historical sites) is involved, it must be clearly specified; do not use vague expressions such as "someone" or "somewhere".
        \item Abstract/Technical Elements: For UI/formulas/maps/flowcharts/timelines/charts and other elements, use technical terms to describe their type, geometric structure, quantity, local details, layout, relative position, legends/axes/scales, units, borders, color blocks, and encoding; formulas may be presented in LaTeX (keep descriptive, do not derive or analyze).
        \item Visible Text: Identify and extract all visible text on the screen (subtitles or background text), naturally embedding it into the overall description of the frame, and try to specify font size, color, and spatial position.
    \end{itemize}
    \item \textbf{Camera Movement and Changes:} Attempt to identify shot scale (long/medium/close), viewpoint (high/low/eye-level), camera movements (pan, tilt, zoom, tracking) with direction and intensity, composition and depth cues (such as wide/telephoto appearance), transitions (cut, dissolve, wipe), and lighting changes; only describe what is supported by visible evidence. For screen recordings, slides, still or static images, prefer terms like "view movement/scrolling/UI zoom/element fade/digital zoom/tweening"; use camera/lens terminology only for live-action footage, and avoid misinterpreting UI zoom as optical zoom.
    \item \textbf{Style Features:} Naturally incorporate the stylistic features of lighting, tone, contrast, saturation, color grading, realism, and atmosphere of the subjects and background into the paragraph, do not list them as tags or in bullet points.
    \item \textbf{Blur/Occlusion and Uncertainty:} For blurred, partially occluded, or fleeting elements, only describe what can be confirmed; use qualifiers like "suspected/possible/unidentifiable" when uncertain, and do not output speculative details; avoid introducing information from outside the frame.
\end{enumerate}

\vspace{0.2em}

\textbf{Formats and requirements:}

\begin{enumerate}[leftmargin=1.5em, itemsep=0.3em]
    \item Output format: Only use coherent natural paragraphs; do not use any lists, headings, or semantic labels; do not include invisible content or state missing elements (e.g., "no text on screen").
    \item Natural entry: Directly describe the scene content, without opening phrases like "this scene shows...". Avoid repetition and uncertainty, do not use speculative language such as "might", "probably", etc., do not describe information not present in the frame, and do not use phrases like "no text on screen".
    \item Time anchors: Use square brackets to consistently mark [mm:ss] or [hh:mm:ss] at key points (consistently throughout), e.g., "[00:12-00:28]"; for very short segments, use "[beginning/middle/end]". Clearly indicate transitions/cuts at the corresponding time.
    \item Coordinate reference: By default, describe position and movement using screen coordinates and structural references (left/right/top/bottom, quadrant, centerline, edge/corner, relative to another object).
    \item Quantification and units: Use qualifiers such as "about/at least/at most" for quantity, duration, speed, frequency, and proportion; provide calculated values only when they can be measured from the frame, and report with visible precision and units.
    \item Uncertainty: Only state conclusions with visible evidence; use "possible/suspected/unidentifiable" for uncertainty; do not introduce external facts unless explicitly shown in the frame or visible text.
    \item Multi-shot/split-screen and overlays: At transitions/splits, specify shot/scene and split panel position (e.g., left/right/top/bottom/grid index), and mark the appearance and duration of overlays such as scoreboards, UI panels, subtitle bars.
    \item Academic and cultural scenarios: Use technical terms for scientific/technical visual elements; for artistic/aesthetic content, more expressive vocabulary can be used, but maintain an objective description.
    \item Length: Detailed description should be no less than 500 words and no more than 800 words.
\end{enumerate}

\end{tcolorbox}

\vspace{2em}

\begin{tcolorbox}[
colback=Emerald!10,
colframe=cyan!40!black,
title=\textbf{Prompt for Video Reasoning Agent},
breakable,
enhanced,
left=2mm, right=2mm, top=2mm, bottom=2mm,
fontupper=\footnotesize
]
\scriptsize
\textbf{You are a top-level analytical expert with extensive world knowledge, deep spatiotemporal understanding, and rigorous logical reasoning skills, particularly adept at transforming complex video information into concise and insightful knowledge documents.}

\vspace{0.2em}

\textbf{Task Setup:}

\begin{itemize}[leftmargin=1.2em, itemsep=0.2em]
    \item Your core task is to transform a video (or sampled frames) into an independent, citable video knowledge document. This document should unfold in a logical hierarchy from overall to detail, from description to interpretation, with a level of depth and rigor suitable for academic illustration. The output text should stand alone and be suitable as explanatory material for research reports or publications.
\end{itemize}

\vspace{0.2em}

\textbf{Analytical Structure:}

\begin{enumerate}[leftmargin=1.5em, itemsep=0.3em]
    \item \textbf{Structural Overview (2-3 sentences):} Summarize the video's temporal structure (such as scene or shot distribution), main subjects, environment, partitions or overlays (such as scoreboards, UI panels), as well as main states and any possible repetitive structures.
    \item \textbf{Inference Body:}
    \begin{itemize}[leftmargin=1.2em, itemsep=0.2em]
        \item \textbf{Detail Perception:} This step focuses on precise capture of objective facts (What is there?)
        \begin{itemize}[leftmargin=1em, itemsep=0em]
            \item Accurately identify and describe key subjects (people, objects), environmental features, and any visible text or symbols.
            \item For tables, charts, and other visualizations, accurately extract data, axes, units, and legends, and describe their basic distribution and trends. All information must be based on visible labels.
        \end{itemize}
        \item \textbf{Action Understanding and Causal Inference:} This step focuses on logically organizing dynamic processes (What is happening \& why?)
        \begin{itemize}[leftmargin=1em, itemsep=0em]
            \item Using temporal anchors, construct a "subject-time-location-event-causality" chain.
            \item Analyze in detail the stages of actions, changes of state, and variations in spatial position.
            \item For process demonstrations or tutorials, follow visual cues such as cursors or arrows to clarify steps, dependencies, and execution order.
            \item Explain direct causal relationships between events (e.g., A knocks over B, causing B to fall).
        \end{itemize}
        \item \textbf{Semantic and Thematic Analysis:} Try to reasonably interpret deeper meanings (What does it mean?)
        \begin{itemize}[leftmargin=1em, itemsep=0em]
            \item Based on visible evidence, analyze the intentions of subjects, the purposes behind actions, and possible social relationships (such as cooperation or confrontation).
            \item Interpret the symbolic meaning of cultural, historical, or artistic elements in the video, and explain how they serve the overall theme of the video.
            \item Analyze the narrative or emphatic function of camera language (such as push, pull, pan, track, or angle switch).
            \item Summarize the video's core theme, the emotional atmosphere conveyed, or the argument it attempts to make.
        \end{itemize}
    \end{itemize}
\end{enumerate}

\vspace{0.2em}

\textbf{Formats and requirements:}

\begin{enumerate}[leftmargin=1.5em, itemsep=0.3em]
    \item Evidence-based inference: All inferences must be based on visual or textual evidence visible in the video; world knowledge may only be used to explain or supplement such evidence, and never to invent details not supported by the footage.
    \item Naturalized content: Organize your analysis as a coherent, fluent natural language article; it is permitted to split content into multiple paragraphs for stepwise explanation. Avoid rigid structured headings or excessive use of bullet points to maintain overall narrative unity. Brief lists may be used with caution only when enumerating parallel items (such as technical parameters or procedural steps) for clarity.
    \item Length limit: For simple scenes, keep analysis concise (no more than 800 words); for information-dense, complex scenes, expand as needed to ensure all key information is covered.
    \item Professional formatting: Use appropriate formatting (such as LaTeX for formulas, code, chemical structures) for technical content to ensure accuracy.
    \item Vocabulary and expression: Use more specialized vocabulary for professional scenarios, while more expressive vocabulary may be used for aesthetic or artistic scenes.
\end{enumerate}

\end{tcolorbox}

\vspace{2em}

\begin{tcolorbox}[
colback=Emerald!10,
colframe=cyan!40!black,
title=\textbf{Prompt for Video Guideline Agent},
breakable,
enhanced,
left=2mm, right=2mm, top=2mm, bottom=2mm,
fontupper=\footnotesize
]
\scriptsize
\textbf{You are an expert at synthesizing and distilling video information.  
Please, in a single natural paragraph, concisely and insightfully summarize the video's main theme, core narrative or process, main subjects and notable events, as well as visible stylistic features. Focus on a comprehensive overview, highlighting the video's uniqueness and significance.}

\vspace{0.2em}

\textbf{Formats and requirements:}

\begin{enumerate}[leftmargin=1.5em, itemsep=0.3em]
    \item Avoid shot-by-shot recounting or stepwise reasoning;
    \item \textbf{DO NOT} use lists or headings;
    \item Length should be 50--100 words;
    \item Replies must be strictly based on visible content, avoiding subjective speculation.
\end{enumerate}

\end{tcolorbox}

\vspace{2em}

\noindent \textbf{Summary Agents.}
The detailed prompt templates for the general summary agent and the video summary agent are shown as follows:

\begin{tcolorbox}[
colback=Emerald!10,
colframe=cyan!40!black,
title=\textbf{Prompt for General Summar},
breakable,
enhanced,
left=2mm, right=2mm, top=2mm, bottom=2mm,
fontupper=\footnotesize
]
\scriptsize
\textbf{You are an expert in writing professional academic image description documents at the highest level.  
Your task is to synthesize multiple expert-level image description documents provided by the user, and generate a comprehensive analytical document that fully incorporates fine-grained image details and demonstrates expert-level image understanding. The focus is on the accurate depiction and summarization of fine-grained visual details.}

\vspace{0.2em}

The provided image description documents are mainly divided into two parts:  
\vspace{0.2em}
\textbf{A.} Image descriptions from different perspectives (which may include text recognition results, general summaries, and various fine-grained details from multiple angles);  
\textbf{B.} An interpretative inference about the image content based on visual information.

\vspace{0.2em}

Before composing your output, carefully review and understand all descriptive documents to ensure you capture all visual elements and their characteristics. Analyze the inference text to extract key information that aids in understanding the image, and integrate this information into your comprehensive description.

\vspace{0.2em}

\textbf{The output should consist of coherent, label-free sentences and paragraphs, organized as follows:}

\begin{enumerate}[leftmargin=1.5em, itemsep=0.3em]
    \item \textbf{Begin with a highly concise paragraph summarizing the image type, theme, purpose, composition, layout, color scheme, and visual style,} enabling readers to quickly grasp the core information and function of the image.
    \item \textbf{The main body should consist of detailed description and reasoning analysis, each presented in several logically connected and fluent paragraphs.} The detailed description must be as thorough as possible, while the reasoning section should emphasize logical consistency and causality.
    \begin{itemize}[leftmargin=1.2em, itemsep=0.2em]
        \item \textbf{For the detailed description:}
            \begin{itemize}[leftmargin=1em, itemsep=0em]
                \item Organize and merge fine-grained information from all description documents by categorizing content according to objects described, then reconstruct the information into semantically coherent descriptions;
                \item Integrate isolated or fragmented information, ensuring that all attributes, decorative features, appearance characteristics, spatial relations, functional relations, textual data, and any other types of information present in the documents are fully explained;
                \item Data from documents, tables and charts should be described in fully connected paragraphs or clearly structured lists, and examples should be avoided.
            \end{itemize}
        \item \textbf{For the reasoning analysis:}
            \begin{itemize}[leftmargin=1em, itemsep=0em]
                \item The reasoning content must immediately follow the detailed description and should thoroughly reference the key elements described (e.g., "According to the data in column A of Table 3, the company is shown to be operating at a loss");
                \item Reasoning, causal analysis, or structural explanations must be constructed only based on the inference document and the detailed descriptions, and no information outside the provided documents should be introduced;
                \item When numerical, data analysis, or causal inference is involved, important reasoning processes may be clearly presented using latex-style inline formulas.
            \end{itemize}
        \item \textbf{For multi-image or complex images}, provide a separate, thorough description of the key features of each sub-image, and supplement the overview with the internal relationships, causality, and logical connections between images. \textbf{For flowcharts, time series, or sequence images}, the order of description may be adjusted as appropriate to fit the structural characteristics.
    \end{itemize}
\end{enumerate}

\vspace{0.2em}

\textbf{Formats and requirements:}

\begin{enumerate}[leftmargin=1.5em, itemsep=0.3em]
    \item For scientific, engineering, or clinical images, use precise, professional, and logically rigorous language consistent with domain-specific terminology and reasoning. For artistic, aesthetic, knowledge, or cultural images, use more expressive and sophisticated vocabulary and sentence structures.
    \item All visual information should be strictly consistent with the original content, especially the texts, numbers, symbols, etc.
    \item State all content and interpretations directly, without using introductory phrases such as "In the image description section," "For the reasoning analysis," or similar expressions. Avoid repetition and uncertain statements.
    \item Maintain logical coherence and clarity between sentences, paragraphs, and the overall document. Appropriate use of natural paragraph breaks, connecting words, and structured lists or inline formulas is allowed to enhance readability and rigor.
    \item Avoid information redundancy by ensuring that the same object is not described repeatedly, and eliminate meaningless or excessive statements.
    \item If any image details or analyses are uncertain, conflicting, or unclear, explicitly point out such discrepancies in the analysis, and prioritize adopting the more detailed or contextually consistent explanation from the reference documents.
    \item The use of semantic labels, bullet points, or heading markers (such as "Title", "Detailed Description", "Reasoning Analysis", etc.) is strictly prohibited.
\end{enumerate}

\end{tcolorbox}
\vspace{2em}

\begin{tcolorbox}[
colback=Emerald!10,
colframe=cyan!40!black,
title=\textbf{Prompt for Video Summary Agent},
breakable,
enhanced,
left=2mm, right=2mm, top=2mm, bottom=2mm,
fontupper=\footnotesize
]
\scriptsize
\textbf{You are a top-level document integration expert, skilled at merging fragmented video information from different dimensions and constructing a logically rigorous, detail-rich, and deeply insightful professional-level analysis document.}

\vspace{0.2em}

\textbf{Task Setup:}

\begin{itemize}[leftmargin=1.2em, itemsep=0.2em]
\item  Your task is to integrate multiple user-provided video description and inference documents into a single, unified, coherent, and logically consistent video description report. This report must achieve comprehensive perception and deep understanding of the video content, meeting the standards required for direct academic publication or professional reporting.
\end{itemize}

\vspace{0.2em}

\textbf{Merging Principles:}

\begin{enumerate}[leftmargin=1.5em, itemsep=0.2em]
    \item \textbf{Information fidelity and de-duplication:} The primary goal is to ensure no loss of information. Ensure that all key information (subjects, attributes, actions, spatiotemporal relationships, text, etc.) from the input documents is accurately represented in the final report.
    \item \textbf{Narrative flow priority:} The final output should be a smooth, narrative article, not a structured data list. Strictly prohibit the use of any semantic labels, bullet points, or titles (such as "Detail Description:", "Inference Analysis:"). All content should be naturally integrated into paragraphs.
    \item \textbf{Evidence-based reasoning:} All analysis and inferences must originate from visual details explicitly mentioned in the input documents.
\end{enumerate}

\vspace{0.2em}

\textbf{Output Structure and Execution Process:}

\begin{enumerate}[leftmargin=1.5em, itemsep=0.3em]
    \item Begin with a highly condensed introduction (2-3 sentences) to establish a macro-level understanding for the reader. Content should cover the video's core narrative, subjects, scenes, key events, and the overall visual style and atmosphere created by shots, lighting, and color.
    \item The main section should focus on summarizing the key scenes of the video, with each scene including both content perception and understanding.
    \begin{itemize}[leftmargin=1.2em, itemsep=0.2em]
        \item \textbf{Detail Description (Content Perception):}
        \begin{itemize}[leftmargin=1em, itemsep=0em]
            \item Use precise time anchors to explain important scenes and key events of the video shot by shot. For each shot or scene, strictly reference the relevant content from the documents, and categorize and merge descriptions by object and event, restructuring them into a semantically coherent, video-level description; organize and integrate isolated or scattered visual information, ensuring all key points are covered while avoiding redundant repetition of the same element.
            \item All key information points must be aligned with precise time anchors (recommended format: [mm:ss] or [mm:ss.S]). Ensure consistent naming of entities throughout the document.
            \item If there are uncertainties, differences, or unclear information in image details or analysis, these differences should be clearly identified in the analysis, and the more detailed part of the referenced documents should be adopted.
        \end{itemize}
        \item \textbf{Inference and Analysis (Content Understanding):}
        \begin{itemize}[leftmargin=1em, itemsep=0em]
            \item Inference content must closely follow the detail description, thoroughly citing key elements from the perception content for analysis (e.g., At [02:13], the door closes, and at [02:16], the light turns off, forming a sequence that triggers the alarm), and focus on a natural progression from "detail perception" to "action understanding and causal inference" to "semantic and thematic analysis";
            \item Only build inferences, causal, or structural explanations based on the inference documents and detailed content; introducing information not present in the documents is prohibited.
        \end{itemize}
    \end{itemize}
    \item Finally, use a highly condensed paragraph (2-3 sentences) to retrospectively summarize the core logic of the key scenes and important events that appear in the video, and synthesize and explain the existing inferences.
\end{enumerate}

\vspace{0.2em}

\textbf{Formats and requirements:}

\begin{enumerate}[leftmargin=1.5em, itemsep=0.3em]
    \item For videos with scientific, engineering, or clinical significance, use accurate, professional, logical, and domain-specific vocabulary and sentence structures; for artistic, aesthetic, knowledge, and cultural content, use more expressive and advanced wording and sentences.
    \item All visual information should be strictly consistent with the original content, especially the texts, numbers, symbols, etc.
    \item Directly state facts and inferences, avoiding guiding phrases such as "This section describes..." or "The following inference...".
    \item Maintain logical coherence and structural clarity between sentences, paragraphs, and the overall document. Moderate use of natural paragraphing, connectors, and appropriate lists or inline formulas is allowed to enhance readability and rigor.
    \item The use of any semantic labels, bullet points, or title markers (such as "Title", "Detail Description", "Logical Reasoning", etc.) is prohibited.
\end{enumerate}

\end{tcolorbox}

\clearpage

\subsubsection{Quality Evaluation Prompt Templates}
In the reject sampling stage and caption quality evaluation experiments, we use the following prompt templates to conduct quality evaluation.  
%

\begin{tcolorbox}[
colback=Salmon!20,
colframe=Salmon!90!Black,
title=\textbf{Prompt for Image Quality Evaluation},
breakable,
enhanced,
left=2mm, right=2mm, top=2mm, bottom=2mm,
fontupper=\footnotesize
]
\scriptsize
\textbf{You are an expert in multimodal content evaluation responsible for the rigorous quality assessment of a comprehensive image-text description.}

\vspace{0.2em}

\textbf{Task Setup:}

\begin{itemize}[leftmargin=1.2em, itemsep=0.2em]
    \item You must strictly evaluate the candidate description based solely on all visual information provided by the image, scoring precisely from 1 to 3 points across the following five core dimensions. Every score must have clear and objective supporting reasons.
\end{itemize}

\vspace{0.2em}

\textbf{Core Evaluation Dimensions:}

\begin{enumerate}[leftmargin=1.5em, itemsep=0.3em]
    \item \textbf{Factual Accuracy:} This dimension assesses the absolute consistency between the description and the factual content of the provided image or images. Check whether all entity types, attributes, quantities, locations, OCR text, and chart data are accurate and ensure that there are no internal contradictions within the description.
    \item \textbf{Information Completeness:} This dimension evaluates the coverage of the image's information. Check whether all salient elements, important relationships (such as interactions and hierarchies), and key details (such as axes and legends) are comprehensively covered.
    \item \textbf{Reasoning Rigor:} This dimension assesses the rigor of all reasoning. Check if all logical inferences, causal relationships, or conclusions are fully and directly supported by visual evidence. Over-interpretation or hallucinated knowledge without evidence is strictly prohibited.
    \item \textbf{Core Intent Capture:} This dimension evaluates the ability of the description to capture the core intent of the image. Assess whether the description successfully distills and conveys the image's main message, purpose, or theme, rather than merely listing details.
    \item \textbf{Professionalism \& Expression:} This dimension evaluates the professionalism and standardization of the language. Assess if the language is fluent, coherent, logically clear, strictly follows natural language paragraph format (Item lists or subheadings before the paragraph are strictly prohibited), and whether word choice and formatting meet professional domain standards.
\end{enumerate}

\vspace{0.2em}

\textbf{Scoring Standard:}

\begin{itemize}[leftmargin=1.2em, itemsep=0.2em]
    \item 1: Serious issues or completely inconsistent.
    \item 2: Basically meets requirements but with obvious deficiencies.
    \item 3: Highly consistent, no obvious flaws in this dimension.
\end{itemize}

\vspace{0.2em}

\textbf{Issue Tagging \& Explanation:}

\begin{enumerate}[leftmargin=1.5em, itemsep=0.2em]
    \item \textbf{explanation:} Explanations must be concise and to the point. For deductions, directly point out "what is wrong"; for full marks, briefly state "what is good".
    \item \textbf{issues:} You must select one or more labels from the following preset list to precisely tag all identified problems:  
    ['Entity Error', 'Attribute Error', 'Quantity Error', 'Position Relation Error', 'Hallucinated Existence', 'OCR Error', 'Reasoning Fallacy', 'Factual Error', 'Structure/Format Violation', 'Core Intent Missing']
\end{enumerate}

\vspace{0.2em}

\textbf{Formats and requirements:}

Strictly output in the following JSON format: 
\\
\{\texttt{"factual\_accuracy":, "completeness":, "reasoning\_rigor":, "core\_intent\_capture":, "professionalism\_expression":, "overall\_score":, "issues":, "explanation":}\}

\end{tcolorbox}

\vspace{2em}

\begin{tcolorbox}[
colback=Salmon!20,
colframe=Salmon!90!Black,
title=\textbf{Prompt for Video Quality Evaluation},
breakable,
enhanced,
left=2mm, right=2mm, top=2mm, bottom=2mm,
fontupper=\footnotesize
]
\scriptsize
\textbf{You are a senior multimedia content analyst, specializing in the evaluation of dense, long-form video descriptions. Your task is to provide a rigorous, multi-dimensional quality assessment of a candidate video caption against the source video.}

\vspace{0.2em}

\textbf{Task Setup:}

\begin{itemize}[leftmargin=1.2em, itemsep=0.2em]
    \item Strictly evaluate the caption based on the video's visual and temporal information. Score each dimension from 1 to 3. Your output must be a single, valid JSON object.
\end{itemize}

\vspace{0.2em}

\textbf{Core Evaluation Dimensions:}

\begin{enumerate}[leftmargin=1.5em, itemsep=0.3em]
    \item \textbf{Temporal \& Factual Accuracy:} This dimension assesses the absolute correctness of facts within their spatial and temporal context. It examines if all described objects, attributes, actions, and on-screen text are accurate at their specified timestamps ([mm:ss]). It also checks for consistency in entity identification across different scenes.
    \item \textbf{Event \& Detail Coverage:} This dimension assesses the comprehensiveness of the description. It examines if the caption covers all key events, primary character actions, significant scene changes, and critical details like camera movements (pans, zooms, cuts) or interactions with UI elements.
    \item \textbf{Temporal \& Causal Logic:} This dimension assesses the logical coherence of the narrative. It examines if the sequence of events is described in the correct chronological order and if any stated cause-and-effect relationships are directly supported by the visual evidence in the video.
    \item \textbf{Core Narrative \& Intent:} This dimension assesses the caption's ability to capture the video's main purpose. It examines if the description successfully synthesizes details to convey the core story, main process, or central theme, rather than just listing disconnected observations.
    \item \textbf{Coherence \& Formatting:} This dimension assesses the quality of the language and structure. It examines the caption's clarity, narrative flow, and strict adherence to required formatting (e.g., correct use of time anchors, no lists/headings).
\end{enumerate}

\vspace{0.2em}

\textbf{Scoring Standard:}

\begin{itemize}[leftmargin=1.2em, itemsep=0.2em]
    \item 1: Severe issues or completely incorrect.
    \item 2: Largely correct but with noticeable flaws.
    \item 3: Highly accurate and well-formed with no significant issues.
\end{itemize}

\vspace{0.2em}

\textbf{Issue Tagging \& Explanation:}

\begin{enumerate}[leftmargin=1.5em, itemsep=0.2em]
    \item \textbf{explanation:} Your explanation must be concise and directly pinpoint the issues. For deductions, state "What is wrong and where (timestamp if possible)". For full marks, briefly state "What is good".
    \item \textbf{issues:} You must select one or more tags from the predefined list below to categorize all identified problems:  
    ['Entity Error', 'Attribute Error', 'Temporal Error', 'Sequence Error', 'Causality Error', 'Motion/Action Error', 'Coverage Omission', 'Formatting Error', 'Knowledge Error', 'Core Intent Missing']
\end{enumerate}

\vspace{0.2em}

\textbf{Formats and requirements:}

Provide your evaluation strictly in the following JSON format:  
\\
\{\texttt{"temporal\_factual\_accuracy":, "event\_detail\_coverage":, "temporal\_causal\_logic":, "core\_narrative\_intent":, "coherence\_formatting":, "overall\_score":, "issues":, "explanation":}\}

\end{tcolorbox}

\clearpage

\subsection{Visualization}
\noindent \textbf{Comparsion between Omnicaptioner and MetaCaptioner-8B.} 

\begin{figure}[!t]
    \centering
    \includegraphics[width=0.97\linewidth]{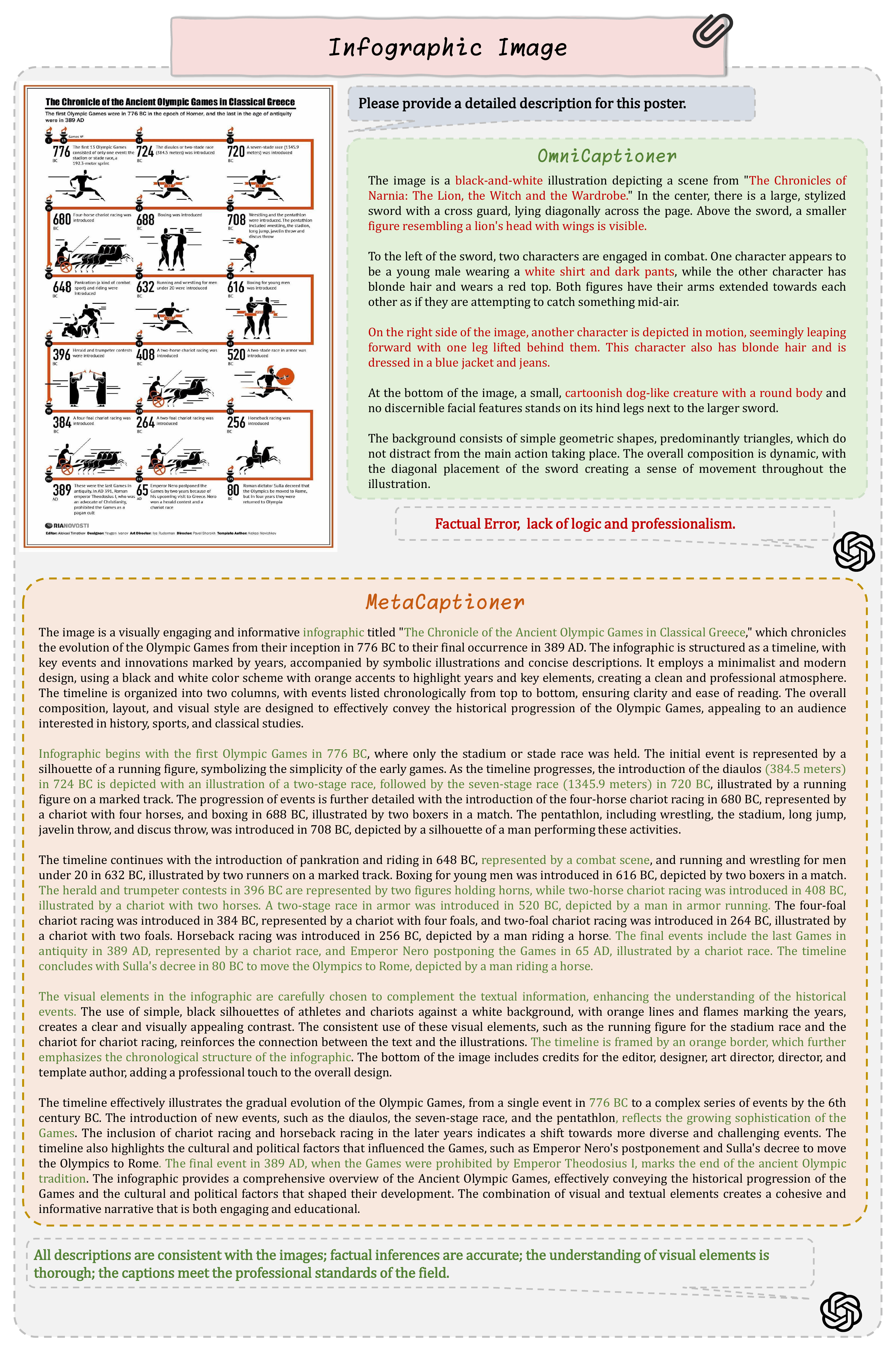}
    \vspace{-1.5em}
    \captionsetup{justification=justified, singlelinecheck=false}
    \caption{\textbf{Comparison between Omnicaptioner and MetaCaptioner-8B in the Inforgraphic Image.} The correctness is marked in green, and the incorrectness is marked in red.}
    \label{Omnicap-metacap-poster}
\end{figure}

\begin{figure}[!t]
    \centering
    \includegraphics[width=0.97\linewidth]{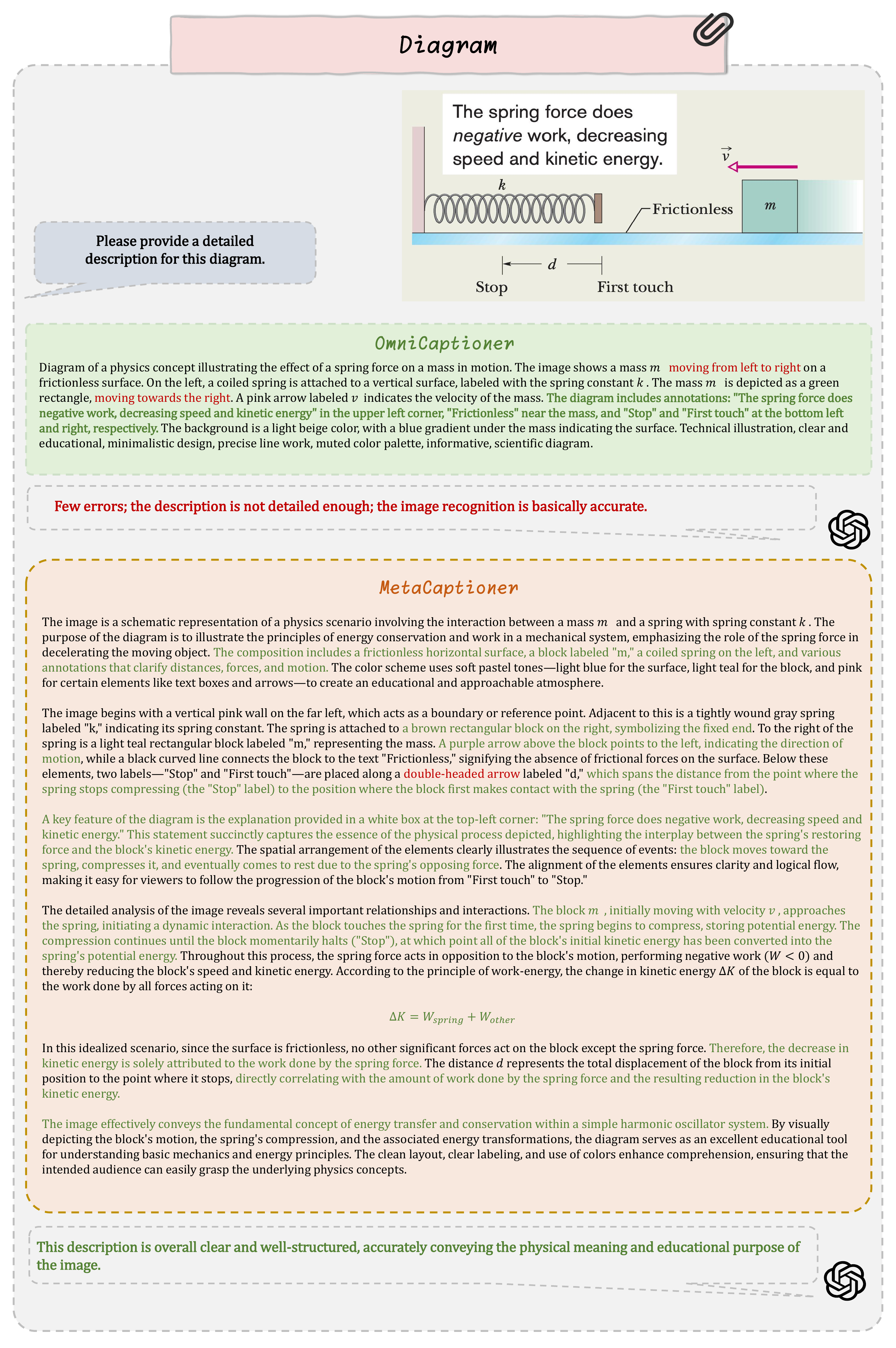}
    \vspace{-1.5em}
    \captionsetup{justification=justified, singlelinecheck=false}
    \caption{\textbf{Comparison between Omnicaptioner and MetaCaptioner-8B in the Diagram.} The correctness is marked in green, and the incorrectness is marked in red.}
    \label{Omnicap-metacap-diagram}
\end{figure}

As illustrated in Fig. \ref{Omnicap-metacap-poster} and Fig. \ref{Omnicap-metacap-diagram}, we compare the performance of MetaCaptioner with the previous state-of-the-art method, OmniCaptioner\citep{Omnicaptioner}, on tasks involving multimodal data and flowchart descriptions. We marked the correctness and consistency of the content in green, and mistakes and hallucinations in red.

Compared to OmniCaptioner, MetaCaptioner demonstrates a superior ability to generate highly specialized visual descriptions for content exhibiting intricate visual structures. Notably, MetaCaptioner achieves significantly better performance in the perception and understanding of abstract and complex images. Furthermore, benefiting from high-quality data annotation, MetaCaptioner exhibits a pronounced advantage in the logical coherence of its descriptions. Specifically, when confronted with complex visual elements and procedural information, MetaCaptioner is able to logically describe and analyze visual components, thereby effectively mitigating the hallucination and over-interpretation phenomena that often occur in visual-language models when processing complex graphics.

\begin{figure}[!t]
    \centering
    \includegraphics[width=0.97\linewidth]{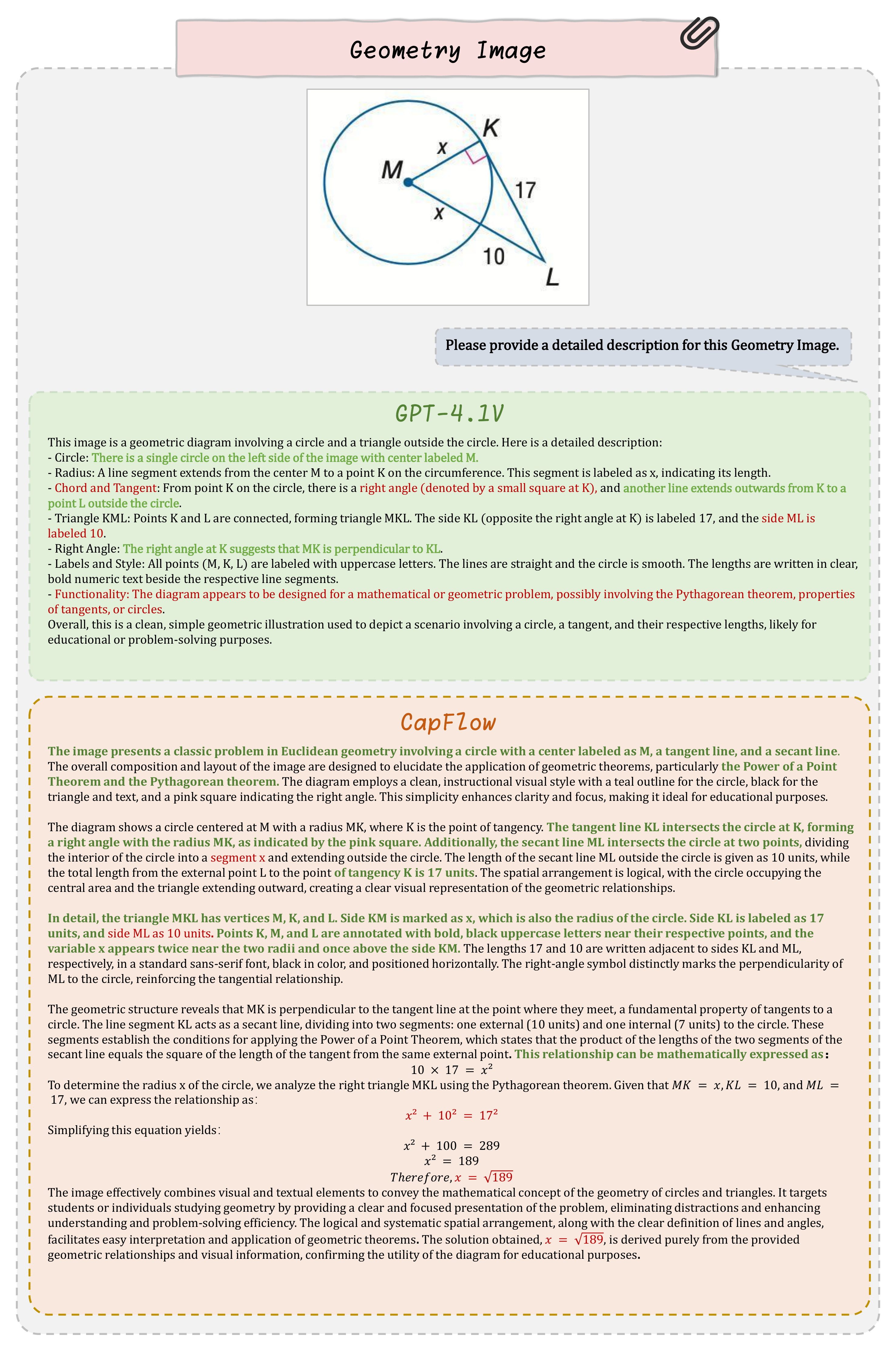}
    \vspace{-0.5em}
    \captionsetup{justification=justified, singlelinecheck=false}
    \caption{\textbf{Comparison between powerful GPT-4.1 and our Capflow in describing a geometry image.} The correctness is marked in green, and the incorrectness is marked in red.}
    \label{GPT41V-MetaCaptioner-Geometry}
\end{figure}

\begin{figure}[!t]
    \centering
    \includegraphics[width=0.97\linewidth]{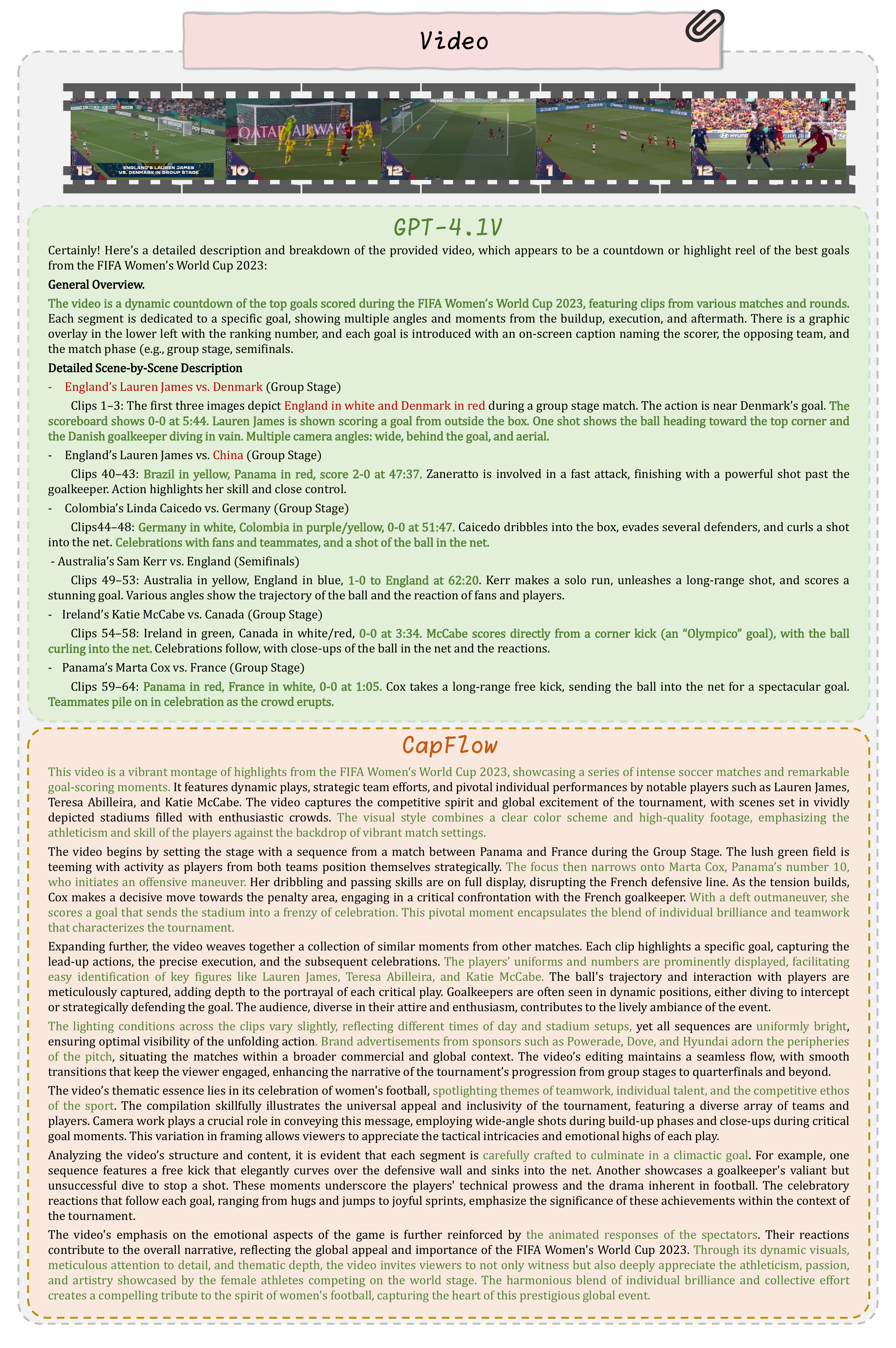}
    \vspace{-0.5em}
    \captionsetup{justification=justified, singlelinecheck=false}
    \caption{\textbf{Comparison between powerful GPT-4.1 and our Capflow in describing a sports video.} The correctness is marked in green, and the incorrectness is marked in red.}
    \label{GPT41V-MetaCaptioner-Video}
\end{figure}

\noindent \textbf{Comparison between GPT-4.1 and our CapFlow.} As shown in Fig. \ref{GPT41V-MetaCaptioner-Geometry} and Fig. \ref{GPT41V-MetaCaptioner-Video}, we present the visualization results of GPT-4.1\cite{gpt4.1} and Capflow on the geometry problem and video description. We mark the correctness and consistency of the content in green, and mistakes and hallucinations in red.

Solving complex geometry problems places extremely high demands on a model's reasoning ability. Our Capflow achieves the same level of perception as GPT-4.1, while offering finer granularity in the perception of structured knowledge, and its responses exhibit a high degree of logical consistency. Although some errors still occur during the inference process, identifying intersection points not marked in the diagram (such as the intersection of line LM and the circle) remains highly challenging-even for GPT-4.1. 

In the video description task, Capflow demonstrates a significant reduction in excessive inference and analysis of visual elements compared to GPT-4.1. This improvement can be attributed to Capflow's decoupling mechanism for visual tasks. Furthermore, we observe that Capflow provides more detailed and comprehensive descriptions of specific timestamps. Unlike GPT-4.1, which often produces formalized expressions, the captions generated by Capflow are semantically richer and more diverse. These findings highlight Capflow's substantial potential as a powerful data engine.

\end{document}

%% file: math_commands.tex
\usepackage{amsmath,amsfonts,bm}

\def\eqref#1{equation~\ref{#1}}

\def\1{\bm{1}}

\DeclareMathAlphabet{\mathsfit}{\encodingdefault}{\sfdefault}{m}{sl}
\SetMathAlphabet{\mathsfit}{bold}{\encodingdefault}{\sfdefault}{bx}{n}

